\documentclass[11 pt]{article}
\usepackage{graphicx}
\usepackage{harpoon}
\usepackage{color}
 \usepackage{stackrel} 
\usepackage{latexsym}
\usepackage{amssymb}
\usepackage{amsmath, bm}
\usepackage{mdframed}
\usepackage{hyperref, url}
\usepackage[show]{ed}
\usepackage{subfigure}
\usepackage[ntheorem]{empheq} 
\usepackage{amsthm, amssymb, amsfonts, latexsym}
\usepackage{lipsum}
\mathchardef\mhyphen="2D

\newcommand{\vertiii}[1]{{\left\vert\kern-0.25ex\left\vert\kern-0.25ex\left\vert #1 
    \right\vert\kern-0.25ex\right\vert\kern-0.25ex\right\vert}}
\newcommand{\vertii}[1]{{\left\vert\kern-0.25ex\left\vert #1 
    \right\vert\kern-0.25ex\right\vert}}

\makeatletter
\DeclareRobustCommand{\cev}[1]{%
  \mathpalette\do@cev{#1}%
}
\newcommand{\do@cev}[2]{%
  \fix@cev{#1}{+}%
  \reflectbox{$\m@th#1\vec{\reflectbox{$\fix@cev{#1}{-}\m@th#1#2\fix@cev{#1}{+}$}}$}%
  \fix@cev{#1}{-}%
}
\newcommand{\fix@cev}[2]{%
  \ifx#1\displaystyle
    \mkern#20mu
  \else
    \ifx#1\textstyle
      \mkern#20mu
    \else
      \ifx#1\scriptstyle
        \mkern#26mu
      \else
        \mkern#26mu
      \fi
    \fi
  \fi
}

\makeatother

\newcommand{\fig}[1]{\epsfbox}
\newcommand{\bp}{\begin{minipage}{3.1cm}}
\newcommand{\ep}{\end{minipage}}
\usepackage{pbox}

\newcommand\blfootnote[1]{%
  \begingroup
  \renewcommand\thefootnote{}\footnote{#1}%
  \addtocounter{footnote}{-1}%
  \endgroup
}

\newtheorem{theorem}{Theorem}[section]
\newtheorem{definition}[theorem]{Definition}
\newtheorem{lemma}[theorem]{Lemma}
\newtheorem{remark}[theorem]{Remark}
\newtheorem{proposition}[theorem]{Proposition}
\newtheorem{corollary}[theorem]{Corollary}
\newtheorem{example}[theorem]{Example}

\newtheorem*{theorem*}{Theorem}

\makeatletter 
\renewcommand{\thefigure}{\@arabic\c@figure}
\makeatother
\usepackage{xr}

\begin{document}

\title{\textbf{\textbf{Transport in Reservoir Computing}}}
\author{G Manjunath$^{1}$ and Juan-Pablo Ortega$^{2}$}
\date{}
\maketitle

\begin{abstract}
Reservoir computing systems are constructed using a driven dynamical system in which external inputs can alter the evolving states of a system. These paradigms are used in information processing, machine learning, and computation. A fundamental question that needs to be addressed in this framework is the statistical relationship between the input and the system states.  This paper provides conditions that guarantee the existence and uniqueness of asymptotically invariant measures for driven systems and shows that their dependence on the input process is continuous when the set of input and output  processes are endowed with the Wasserstein distance. The main tool in these developments is the characterization of those invariant measures as fixed points of naturally defined Foias operators that appear in this context and which have been profusely studied in the paper. Those fixed points are obtained by imposing  a newly introduced stochastic state contractivity on the driven system that is readily verifiable in examples. Stochastic state contractivity can be satisfied by systems that are not state-contractive, which is a need typically evoked to guarantee the echo state property in reservoir computing. As a result, it may actually be satisfied even if the echo state property is not present.
\end{abstract}

\bigskip


\makeatletter
\addtocounter{footnote}{1} \footnotetext{%
Department of Mathematics \& Applied Mathematics, University of Pretoria,
Pretoria 0028, South Africa.
{\texttt{manjunath.gandhi@up.ac.za}}}
\addtocounter{footnote}{1} \footnotetext{%
Division of Mathematical Sciences, 
Nanyang Technological University,
21 Nanyang Link,
Singapore 637371.
{\texttt{Juan-Pablo.Ortega@ntu.edu.sg}}}
\makeatother

 \blfootnote{We thank Lukas Gonon and Lyudmila Grigoryeva for interesting exchanges that inspired some of the results in this paper. GM acknowledges support from the National Research Foundation of South Africa under Grant UID 115940 as well as the hospitality of the Division of Mathematical Sciences of the Nanyang Technological University which funded the visit in September 2022 during which this paper was written. JPO acknowledges partial financial support  coming from the Swiss National Science Foundation (grant number 200021\_175801/1).  }

\section{Introduction} 
\label{Sec_Intro}
  
Transport in dynamical systems is studied at both  microscopic and macroscopical levels. On the one hand, at the microscopic level, if one is interested in the motion of a particle in a fluid, and the particle is assumed to be so light that it can do nothing but follow the liquid, then the motion of the fluid totally determines the fate of the particle. In particular, the different dynamical properties of the fluid can create transport barriers for the particle (see, for instance, \cite{aref2002development, wiggins2013chaotic}) trapping its trajectory in a subset of the phase space. When the motion of individual trajectories is not possible due to the inherent loss of predictability that is typical in very general classes of dynamical systems, transport  at the macroscopic level is useful. In such a macroscopic study, one aims to make predictions regarding the longtime evolution of ensembles of trajectories and the term transport refers to the properties of the time evolution of measures  \cite{lasota1998chaos}. 
Loosely stated, such transport does not describe anymore the motion of a particle but instead describes how mass ``accumulates" over a period of time. In the language of  dynamical systems, such transport concerns  the evolution of an ensemble of initial conditions. Such macroscopic transport can yield very simple measure dynamics even when the underlying microscopic dynamics is very complex. 

\begin{center}\begin{figure}[h!]\centering
\includegraphics[width=16cm]{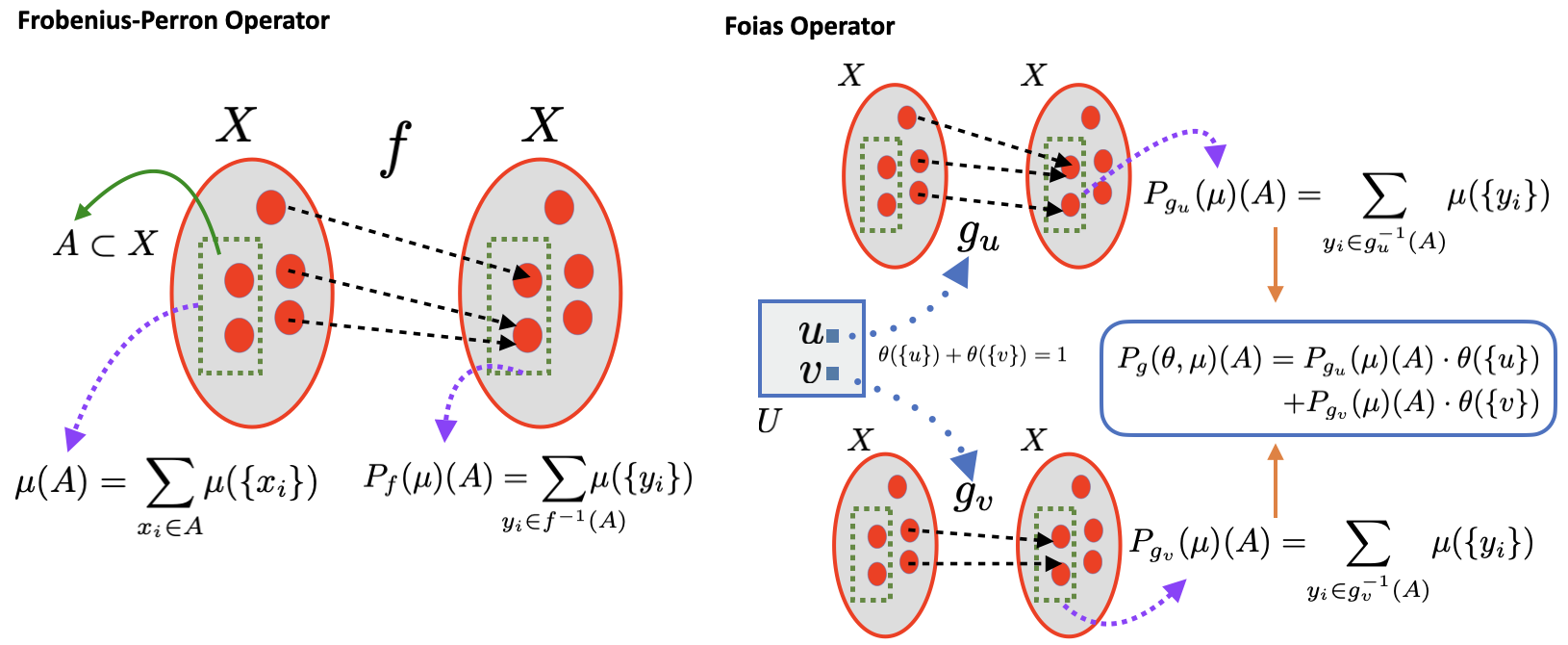}
\caption{An illustration of  Frobenius-Perron and the Foias operators. The Frobenius-Perron operator  pushes forward measures using a dynamical system. The Foias operator is a non-autonomous generalization that integrates the push-forwards given by a driven system with respect to a measure in the input space.} 
\label{Fig_Foias_Operator}
 \end{figure}
\end{center}

An important case takes place when the simplified macroscopic dynamics exhibits a single limit. This feature is related to certain statistical stability which means that a swarm of initial conditions with different initial distributions/densities can all converge (typically in the $L^1$ norm) to an asymptotic distribution/density which is known as an {\bf invariant measure/density} \cite{lasota1998chaos}. The invariant measure/density  sheds light on how the asymptotic states of the system get distributed. 
In the case of time-independent or autonomous systems, the time evolution of measures/densities is well-studied using the  {\bf Frobenius-Perron operator} (see, for instance, \cite{lasota1998chaos} and Figure~\ref{Fig_Foias_Operator}). Non-autonomous extensions of these results  present serious challenges since the input  affects the time-evolution of the measures at each time step. These difficulties are even more pronounced whenever other natural metrics like, for instance, the Wasserstein distance are used in the space of measures. In order to visualize why this is so, recall that for autonomous systems the Wasserstein distance intuitively corresponds to the effort in moving a mount of mass that is not disturbed during transportation; for a nonautonomous system, it would correspond to  the effort of moving a mount of mass that can also be disarranged during transportation.

 In this work, we consider a class of  time-dependent dynamical systems that arise in the field of systems theory and, more specifically, in {\bf reservoir computing (RC)}~\cite{jaeger2001, Jaeger04, maass1, maass2, lukosevicius, tanaka:review}.   RC uses input-output systems that are defined with the help of a {\bf driven} or {\bf state-space system}, that is, a continuous function $g: U \times X \to X$ on an {\bf input space} $U$ and a {\bf state space} $X$ (both are metric spaces). The main difference between general driven systems and RC is that for the latter and for supervised machine learning applications, the function $g$ is not trained but (partially) randomly generated, and the corresponding input-output system is obtained out of a (functionally simple) observation equation of the states. Various families of reservoir computing systems have been shown to exhibit universal approximation properties \cite{RC6, RC7, RC8, RC13, RC20, RC12}.  A {\bf solution} of $g$ for a given bi-infinite input  $\{u_n\}_{n \in \Bbb Z}$ is a bi-infinite sequence $\{x_n\}_{n \in \Bbb Z}$ whenever the equality $x_{n+1}=g(u_n,x_n)$ is satisfied for all $n\in \mathbb{Z}$. The terms $x _n $ of the solution are referred to as state  values or reservoir states in the RC context.

If for each input $\{u_n\}_{n \in \Bbb Z}$ there exists exactly one solution $\{x_n\}_{n \in \Bbb Z}$, then $g$ is said to have the {\bf echo state property (ESP)} \cite{jaeger2001} or the {\bf unique solution property (USP)} \cite{manjunath2022embedding}. Often in practice, only a class of inputs is considered when formulating the ESP, like for instance the one coming from the realizations of a stochastic process. In this work, we place ourselves in a setup that does not necessarily imply the ESP for all possible inputs, as it has been empirically demonstrated, for instance across the echo state networks (ESNs) literature~\cite{jaeger2001, Jaeger04}, that the performance of ESNs is sometimes enhanced when the reservoir dynamics does not have this property \cite{lukosevicius}.  
 
From a qualitative dynamics point of view, the unique solution property is equivalent (at least for compactly driven systems) to the fact that for repeated runs of the RC system with a given input sequence and using different initial conditions in the state space, the resulting state sequences get closer and closer to a solution when the RC runs are longer and longer. This property is called the {\bf uniform attracting property (UAP)} \cite[Definition 3]{manjunath2022embedding} and amounts to the system exhibiting what is called  a {\bf uniform point attractor}.  In \cite[Theorem 1]{manjunath2022embedding} it has been shown that the UAP is equivalent to the USP. 

In order to develop some intuition about these facts, we will use as an example echo state networks (ESNs), a family of RC systems that has been profusely used in applications. ESNs are determined by a state map $g: \mathbb{R}^d \times \mathbb{R}^N \to \mathbb{R}^N$ defined as
\begin{equation}
\label{esn for later}
g(u,x) = \overline{\tanh}(Cu + \alpha A x),
\end{equation}
where $A$ and $C$ are called the reservoir (or connectivity) and input matrices, respectively, which have appropriate dimensions and, in the RC context, are randomly generated. The symbol $\overline{\tanh}$ denotes the nonlinear activation function $\tanh$ applied in a component-wise manner.  Notice that when a large-amplitude input is used in this RC system, the activation function $\tanh$ is saturated because the reservoir neurons become highly stimulated, the $\tanh$ quenches strongly and, as a consequence, the initial condition is forgotten. On the other hand, for small-amplitude inputs, such ``washing out" qualities may be lost. For instance, a constant zero input, is a good candidate for the USP to cease to hold. In practice, however, the relevant input range frequently contains zero. More specifically, all that is often mentioned about permissible inputs is their range, which consequently yields non-typical inputs like the constant-zero signal as an allowable input, that must hence be accounted for when determining the USP.  Much has been studied regarding the intricacies associated to the USP and its dynamical implications. For instance,  in \cite{Manjunath:Jaeger}, the USP with respect to an input has been considered  and, in particular, it has been shown that even if the USP with respect to all inputs does not hold, the USP can still hold with probability $1$ when some given  stationary ergodic process is considered as the input. However, vital questions remain unanswered. 
 
This paper studies how measures are transported in the RC framework with the goal of answering the following fundamental questions: {\bf (i)} If input sequences originate from a stationary stochastic process, how would an ensemble of reservoir state sequences with an arbitrary distribution get asymptotically distributed? In particular, is there an invariant distribution/measure available for the RC systems like in the case of a Markov chain?  {\bf (ii)} When such an invariant measure exists, does it depend continuously on the input stationary measure when a natural measure of dissimilarity between probability measures like, for instance, the Wasserstein distance is employed?  The Wasserstein distance is a particularly appropriate choice first because of its good analytical properties but, more importantly, because of its relation with optimal transport (see \cite{villani2009optimal, panaretos2020invitation} and references therein). 

\begin{center}\begin{figure}[h!]\centering
\includegraphics[scale=0.5]{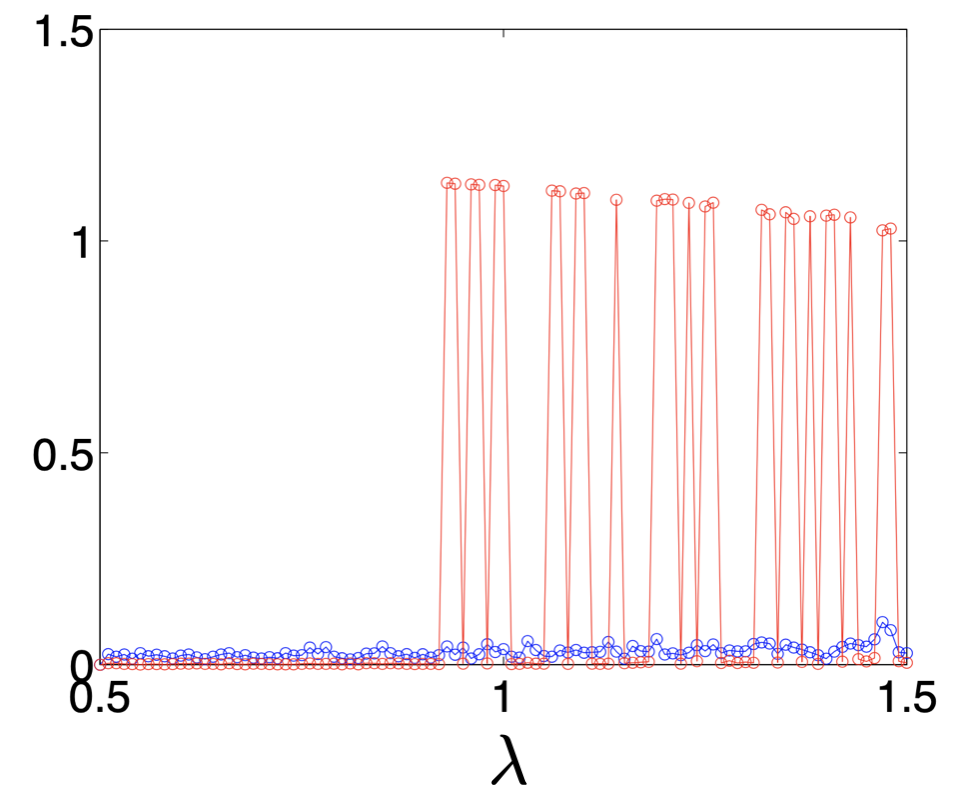}
\caption{Consider an ESN of the type \eqref{esn for later} with input and  reservoir matrices  randomly initialized and with a reservoir matrix set to having a spectral radius of $1.5$. This condition makes the ESN incompatible with the USP \cite{jaeger2001}. Input is fed into the ESN that follows an exponential distribution with parameter $\lambda$. In blue we depict the  Wasserstein-1 distance (introduced later on) between the exponential distributions with parameters $\lambda$ and $\lambda+\epsilon$ (with $\epsilon = 0.01$) against $\lambda$. In red is the corresponding Wasserstein-1 distance observed between the distributions of the states of a single neuron observed.   The distances are calculated using 3500 data points. Larger values on the vertical axis point to the discontinuity of the distribution of the states of the neuron as $\lambda $ is varied.} 
\label{Fig_Exp}
 \end{figure}
\end{center}

These questions are relevant in two contexts: {\bf (a)} Any future notion of stochastic reservoir computing where the stochasticity of the reservoir is controlled by the input requires the properties {\bf (i)} or {\bf (ii)}. {\bf (b)} The echo state property or the USP in the RC framework has been found to yield 
certain stability properties; for instance, the USP guarantees an input-related stability \cite{manjunath:prsl} that implies that close-by input sequences lead to close-by reservoir state sequences. In this context, it is natural to ask what other hypotheses may be required so that close-by input stationary distributions/measures  lead to 
close-by invariant reservoir stationary distributions/measures. At this point, it is important to emphasize that in the absence of the USP, such robustness is actually not available in general. Indeed, as we numerically illustrate in Figure~\ref{Fig_Exp} with a ESN of the type introduced in \eqref{esn for later} that does not have the USP,  large variations in the distributions of the reservoir states can be obtained even when the input distribution is varied slightly. 

In this work we propose to deal with the questions {\bf (i)} and {\bf (ii)} by providing sufficient conditions under which these properties are satisfied. Since we intend to ensure that these features are available even when the USP is not satisfied, 
 we consider a notion in which the resulting reservoir state sequences can  be separated from each other possibly with a with non-vanishing  probability.
More precisely,  given a probability measure $\theta$ on the input space $U$, and  $0<c<1$, we say that the driven system $g$ is a {\bf $(\theta,c)$-stochastic contraction} when
\begin{equation}
\label{stochastic contractivity 1}
	\int_{U} d_{X}(g(u,x),g(u,y)) \, d\theta(u) \le c\, d_X(x,y)\quad \mbox{holds for all $x,y \in X$}. 
\end{equation} 
We stress that stochastic contractivity, a requirement commonly invoked to ensure the echo state property (called USP here) in reservoir computing, can be satisfied by systems that are not state-contractive, and may even be satisfied in the absence of the echo state property (see Remark \ref{stochastic not esp}).
Similar conditions have been formulated in the literature, mostly in the $L ^p $ context to, for example, prove the stability of functional autoregressive models (see \cite[Chapter 6]{Marie:Duflo} and references therein).
In our setup, and since the questions {\bf (i)} and {\bf (ii)} above concern distributions of solution sequences in $X$ rather than just the values in $X$, we shall introduce later on in Definition \ref{eq_G} a system $\mathbf{G}$ in sequence space induced by $g$, and use the above contraction property to handle {\bf (i)} and {\bf (ii)} in Theorems~\ref{Theorem_contraction}  and~\ref{Theorem_im_continuity}. Regarding the time evolution of measures,  in the context of autonomous systems the Frobenius-Perron operator \cite{lasota1998chaos} has been traditionally used. In our non-autonomous setup, we utilize one of its generalizations, namely the the so-called {\bf Foias operator}, which was used, for instance, in \cite[Chapter 12]{lasota1998chaos} to analyze systems driven by IID noise and studied mostly on spaces equipped with the $L ^1 $ norm (see  Figure~\ref{Fig_Foias_Operator} for an illustration).  Also, while transport problems in autonomous dynamical systems are completely solved when the exit times from each measurable set is known \cite{wiggins2013chaotic}, with RC systems such questions have not yet been addressed. In that respect, this paper can be viewed as the first step to answer the question of the convergence towards a stationary distribution or invariant measure and, more generally,  whether we can reliably use the statistical information of the reservoir for information processing.

The general organization of the paper is as follows: Section \ref{Preliminaries} introduces the setup in relation with driven systems and their associated Foias operators. In particular, Proposition \ref{conditions Foias wd} determines when such an operator is well-defined as a map between Wasserstein spaces. Section \ref{Driven systems in sequence spaces} contains a detailed account on how a driven system $g$ naturally induces  another driven system $\mathbf{G} $ in sequence space. The sequence space representation is important since it provides additional tools for the characterization of the solutions of a driven system and of their reachable sets. Section \ref{Stochastic contractions and invariant measures for driven systems} is the core of the paper and contains the main results. More specifically, we prove in this section the {\bf existence and uniqueness of invariant measures for the Foias operators} in both the state (Theorem \ref{Theorem_im_continuity}) and sequence spaces (Theorem \ref{continuous foias fixed point seq}), as well as the {\bf continuity of their dependences on the input process}. 
The main tool to achieve this is Banach's Fixed Point Theorem, which requires two  conditions, namely contractivity and continuity, which will be implied  for the Foias operators in both the state and sequence spaces by {\bf conditions that are readily verifiable and exclusively formulated for the driven system $g$ defined in the state space}. Indeed, most of the developments in that section consist in showing that the contractivity and continuity conditions imposed on $g$ translate to similar properties at the level of the Foias operators in both the state and sequence spaces. The contractivity question is mostly treated in Subsection \ref{Stochastic state contractivity and contractive Foias operators} where it is shown that the newly introduced notion of {\bf stochastic state contractivity} \eqref{stochastic contractivity 1} for the driven system $g$, ensures that the Foias operators  in state  and in sequence spaces are also contractive with respect to the Wasserstein distance (see Figure \ref{Fig_Contraction_Implication} for a summary of the implications between  contractivity in different spaces). We emphasize that  stochastic state contractivity is less restrictive than the standard state contractivity condition evoked to ensure the USP. All the continuity questions are contained in Subsection \ref{Continuity of driven systems and their Foias operators} (see Figure \ref{Fig_Continuity_Implication} for a summary of the different continuity implications). Subsection \ref{The fixed points of the Foias operator} contains the two main results Theorem \ref{Theorem_im_continuity} and Theorem \ref{continuous foias fixed point seq}. Section \ref{Conclusions}  concludes the paper.

\section{Preliminaries}
\label{Preliminaries}
 
\noindent {\bf The setup.} As we already mentioned in the introduction, we place ourselves in the context of {\bf driven systems} induced by a function $g: U \times X \to X$ which has as domain the metric {\bf input  space} $(U,d_U)$ and the metric {\bf state space} $(X,d_X)$. 
We say that a bi-infinite {\bf output sequence} $\mathbf{x}=\{x_n\}_{n\in \mathbb{Z}}  \in X ^{\mathbb{Z}}$ is {\bf compatible} or is a {\bf solution} for the {\bf input sequence} ${\bf u} = \{u_n\}_{n\in \mathbb{Z}} \in U ^{\mathbb{Z}}$ when the following identity is satisfied
\begin{equation} 
\label{eq_DrS}
x_{n+1} = g(u_n, x _n), \quad \mbox{for all $n \in \mathbb{Z}$. } 
\end{equation}
The driven system $g$ has the {\bf unique solution property (USP)} if for each input sequence there is a unique output sequence that is compatible with it.
In the reservoir computing framework, the USP is usually referred to as the echo state property (ESP) and is often ensured by imposing  various contraction properties. The USP guarantees the existence of a unique causal and time-invariant  {\bf  filter} $U _g: U ^{\mathbb{Z}} \rightarrow X ^{\mathbb{Z}} $ (see \cite{Boyd1985} or \cite{RC9} for detailed definitions) which is  characterized by the relation
\begin{equation}
\label{filter with USP}
U _g ({\bf u})_{n+1} = g\left(u_n, U _g ({\bf u})_{n}\right), \quad \mbox{for all $n \in \mathbb{Z}$. } 
\end{equation}
We shall show in our work that in the presence of stochastic inputs,  even if the USP condition is not satisfied by the driven system, we can still use the {\bf Foias operator} to associate to it a {\it continuous input-output system in the space of stochastic processes}. 

\noindent {\bf Wasserstein distances.} We recall some relevant definitions next. Suppose that $(Y , d _Y) $ is a {\bf Polish space} (that is, it is complete and separable) and denote by $P(Y) $ the set of Borel probability measures. Let $p \in [0, \infty) $ and define the {\bf Wasserstein space} $P _p(Y)$ of order $p$ as
\begin{equation*}
P_{p}(Y):=\left\{\mu \in P(Y) \left | \,  \int_{Y}  \right. d_Y(y_{0}, y )^p d\mu(y)<+\infty\right\} ,
\end{equation*}
where $y _0 \in Y $ is arbitrary since it can be shown that this definition does not depend on the point $y _0 $. This space can be made into a Polish space by using the {\bf Wasserstein-p distance} $W_{p}: P_{p}(Y) \times P_{p}(Y) \rightarrow  P_{p}(Y)$ (see \cite[Theorem 6.18]{villani2009optimal})
\begin{equation}
\label{characterization Wasserstein}
\begin{aligned} 
W_{p}(\mu, \nu) &=\left(\inf _{\pi \in \Pi(\mu, \nu)} \int_{Y} d _Y(x, y)^p d \pi(x, y)  \right)^{1/p}\\ 
&=\inf _{\pi \in \Pi(\mu, \nu)} \left\{\left({\rm E}\left[d_Y(X, Y)^p \right]\right)^{1/p}\mid  \operatorname{law}(X)=\mu, \  \operatorname{law}(Y)=\nu\right\}, 
\end{aligned}
\end{equation}
where $\Pi(\mu, \nu)$ is the set of all joint Borel probability measures on $Y \times Y$ whose marginals are $\mu$ and $\nu$, that is, $\mu(A) = \pi (A \times Y)$ and $\nu(A) = \pi (Y \times A)$ for all regular Borel subsets $A\subset Y$. 

All the work in this paper will be conducted using the Wasserstein-1 distance $W _1 $ that in the sequel will be denoted just as $W$. The {\bf Kantorovich-Rubinshtein } duality formula (see \cite{kantorovich1958space, kravchenko2006completeness} or \cite[Theorem 11.8.2]{dudley2018real}) provides an alternative expression for the Wasserstein-1 distance, namely
\begin{equation} 
\label{eqn_Wass1}
W(\mu, \nu) = \sup_{f \in \operatorname{Lip}_1(Y, \mathbb{R})}\bigg\{ \int_Y f d\mu - \int_Y f d \nu  \bigg\},
\end{equation}
where $\operatorname{Lip}_1(Y, \mathbb{R})$ denotes the set of all real-valued functions $f$ on $Y$ so that $|f(y_1)-f(y_2)| \le d_Y(y_1,y_2)$, for all $y _1, y _2\in Y $ or, equivalently, the set of real-valued Lipschitz continuous functions with Lipschitz constant smaller or equal to one.


Since we shall be using profusely  Lipschitz continuous functions, we recall that given two metric spaces $(Y, d_Y)$, $(X, d _X)$, and $ c> 0 $, we define the space $\operatorname{Lip}_c(Y,X)$ of {\bf $c$-Lipschitz continuous} functions between them as
\begin{equation*}
\operatorname{Lip}_c(Y,X)= \left\{f:Y \rightarrow X\mid d _X(f(y _1), f(y _2))\leq c d _Y(y _1, y _2) \mbox{ for all $y _1, y _2 \in Y $ such that  $y _1\neq y _2 $}\right\}.
\end{equation*}
When $Y=X$ we just write $\operatorname{Lip}_c(Y) $. Given  $f:Y \rightarrow X $ we define its {\bf Lipschitz constant} $\left\|f\right\| _{\operatorname{Lip}} $ as 
$${\displaystyle \left\|f\right\| _{\operatorname{Lip}} =\sup \left.\left\{\frac{d _X(f(y _1), f(y _2))}{d _Y(y _1, y _2)} \right| \mbox{$y _1, y _2 \in Y $ such that  $y _1\neq y _2 $}\right\}}.$$
We clearly have that
\begin{equation*}
\operatorname{Lip}_c(Y,X)= \left\{f:Y \rightarrow X\mid \left\|f\right\| _{\operatorname{Lip}}\leq c \right\}.
\end{equation*}
Finally, we denote the set of {\bf Lipschitz continuous} functions by $\operatorname{Lip}(Y, X)$ and we define it with this notation as
\begin{equation*}
\operatorname{Lip}(Y, X)=\left\{f:Y \rightarrow X\mid \left\|f\right\| _{\operatorname{Lip}}< +\infty \right\}.
\end{equation*} 

An important fact that we will use repeatedly is that  $W  $ metrizes the weak convergence in $P_{1}(Y) $ (see \cite[Theorem 6.9]{villani2009optimal}). More specifically, we say that the sequence $\left\{\mu_n\right\}_{n \in \mathbb{N}}$ of measures in $P_{1}(Y) $ {\bf converges weakly} to $\mu\in P_{1}(Y) $, whenever for any continuous real-valued function $f$ such that  $|f(y)| \leq C(1+d(y _0, y)) $, for some $C \in \mathbb{R} $, and some (and then any) $y _0 \in Y $, we have that
\begin{equation}
\label{characterization weak conv}
\int _Y f(y) d\mu _n(y) \rightarrow \int _Y f(y) d\mu (y).
\end{equation}
We say that  $W  $ metrizes the weak convergence in $P_{1}(Y) $ because the statement $\left\{\mu_n\right\}_{n \in \mathbb{N}}$ converges weakly to $\mu\in P_{1}(Y) $ is equivalent to $W(\mu_n, \mu) \rightarrow 0 $. See \cite[Definition 6.8]{villani2009optimal} for other characterizations of weak convergence in $P_{1}(Y) $.

We refer to the Chapter 6 of the monograph \cite{villani2009optimal} for the central role of the Wasserstein metric in the study of optimal transport. 
In particular, unlike the total variation norm or the Kullback-Leibler divergence, also helps in comparing measures that are not absolutely continuous with respect to each other, a situation that often arises in practice while considering empirical distributions.

\medskip

\noindent {\bf The Frobenius-Perron and the Foias operators}. Using the terminology in \cite[Chapter 12]{lasota1998chaos} and in the same setup as in the previous paragraph, we can associate to each Lipschitz continuous discrete-time dynamical system on $Y$ a natural operator $P _f: P _1(Y) \rightarrow P _1(Y)  $ that describes how probability distributions on $Y$ are mapped by the dynamical system. More specifically, let   $f \in \operatorname{Lip}(Y)$ be a Lipschitz continuous self-map of $Y$. The {\bf Frobenius-Perron operator} $P _f: P _1(Y) \rightarrow P _1(Y)  $ associated to $f$ is defined by $P _f(\mu)= f _\ast \mu $, with $ f _\ast \mu $ the pushed-forward measure of $\mu \in P _1(Y)$ by $f$ given by $f _\ast \mu(A)= \mu \left(f ^{-1}(A)\right)$, for any Borel subset $A \subset Y$.

The Lipschitz condition on $f$ implies that $P _f(\mu) \in P _1 (Y)$. Indeed, if $c \in \mathbb{R}  $ is a Lipschitz constant of $f$, then for any $y \in Y $ the triangle inequality implies that
\begin{equation*}
d_Y(y _0, f(y))\leq d_Y(y _0, f (y _0))+ d_Y(f (y _0), f (y))\leq d_Y(y _0, f (y _0))+ c d_Y(y _0,y),
\end{equation*}
which ensures that
\begin{equation*}
 \int_{Y} d_Y\left(y_{0}, y\right) dP _f(\mu)(y)= \int_{Y} d_Y\left(y_{0},f(y)\right) d\mu(y)\leq 
d_Y(y _0, f (y _0))+ c \int_{Y} d_Y(y _0,y)d\mu(y)<+\infty.
\end{equation*}
Notice that {\it if $ d_Y $ is a bounded metric, then the Frobenius-Perron operator $P _f: P _1(Y) \rightarrow P _1(Y)  $ is defined for any measurable self-map} $f$ of $Y$ and {\it if $Y$ is compact then $P _f $ is defined for any continuous map $f$}.

Additionally, if $\operatorname{Lip} _c(Y)$  denotes the space of $c$-Lipschitz continuous dynamical systems on $Y$ and $ f \in \operatorname{Lip} _c(Y)$,  then $P _f \in \operatorname{Lip} _c(P _1(Y))$ when $P _1(Y) $ is endowed with the Wasserstein-1 distance. Indeed, using the  characterization in \eqref{characterization Wasserstein} we have
\begin{multline*}
W(P _f (\mu), P _f (\nu)) =\inf _{\pi \in \Pi\left(P _f (\mu), P _f (\nu)\right)}\int_{Y} d _Y(x, y) d \pi(x, y)\\
\leq \inf _{\pi \in \Pi\left( \mu,  \nu\right)}\int_{Y} d _Y(x, y) d\left( \left(f \times f\right)_\ast \pi\right)(x, y)\leq
\inf _{\pi \in \Pi\left( \mu,  \nu\right)}\int_{Y} d _Y(f(x), f(y)) d\pi(x, y)\\
\leq  \inf _{\pi \in \Pi\left( \mu,  \nu\right)}\int_{Y} c d _Y(x, y) d\pi(x, y)=cW(\mu, \nu), \quad \mbox{for any $\mu, \nu \in P _1(Y)$.} 
\end{multline*}

The notion of  Frobenius-Perron operator for a dynamical system can be extended to driven systems $g $ of the type introduced in \eqref{eq_DrS}, in which case is called the {\bf Foias operator} (see \cite[Definition 12.4.2]{lasota1998chaos}). 

\begin{definition}
\label{def:foias}
Let $g:U \times X \rightarrow X $ be a measurable driven system that has as domain the input $(U, d_U)$ and state $(X, d_X)$ spaces that are assumed to be Polish spaces. The {\bf Foias operator} $P _g: P(U) \times P _1(X) \rightarrow P _1(X) $ associated to $g$ is defined by
\begin{equation}
\label{def:foias:expression}
P _g( \theta, \mu)=\int _U g _{u\, \ast} \mu\, d \theta(u), \  \mbox{where $g _{u}:X \rightarrow X $ is defined by $g _{u}(x)=g(u,x)$, for all $u \in U $.}
\end{equation}
The term measurable means that the preimage $g ^{-1}(A) $ by $g$ of any Borel subset $A \subset X $  of $X$ is a Borel subset of $U \times X $. The equality that defines $P _g( \theta, \mu) \in P _1(X)$ in \eqref{def:foias:expression} is an abbreviation for the measure that for any Borel subset $A \subset X $ takes the value
\begin{equation*}
P _g( \theta, \mu)(A)=\int _X \left( \int _U {\bf 1}_{A}\left(g (u,x)\right)\, d \theta(u) \right)\,  d\mu(x),
\end{equation*}
with ${\bf 1}_{A}: X \rightarrow \left\{0,1\right\} $ the indicator function of $A$.
\end{definition}

\begin{remark}
\label{gstar and pg are the same remark}
\normalfont
It is easy to show using Fubini's Theorem that the Foias operator coincides with the push-forward map $g _\ast: P(U \times X) \longrightarrow P(X)$ when restricted to product measures in $ P(U \times X) $ of the form $\tau(C \times A)= \theta(C) \cdot \mu (A)$, with $C $  and $A $ Borel sets in $U$ and $X$, respectively, and $\theta \in P(U) $, $ \mu \in P_1(U) $. In other words, the Foias operator $P _g $ coincides with the push-forward  map $g _\ast $ of a driven system $g$ when applied to independent random variables in $U$ and $X$ (that is, the ones that have laws $\theta  $ and $\mu$). Indeed, for any such measure and any Borel set $A\subset X$, Fubini's theorem guarantees that:
\begin{multline}
\label{gstar and pg are the same}
g _\ast\tau(A)=\int_{X} {\bf 1}_A(x)\, d \left(g _\ast\tau\right) (x)=\int_{U \times X} {\bf 1}_A(g(u,x))\, d \tau (u,x)=\int_{U \times X} {\bf 1}_A(g(u,x))\, d \theta(u) d \mu(x)\\
=\int_{U \times X} {\bf 1}_A(g(u,x))\, d \theta(u) d \mu(x)=\int _X \left( \int _U {\bf 1}_{A}\left(g (u,x)\right)\, d \theta(u) \right)\,  d\mu(x)= P _g( \theta, \mu)(A). 
\end{multline}
\end{remark}

The following result provides conditions that ensure that the Foias operator is well-defined and that, in particular, maps into $P _1(X) $. Most of the time in this paper we shall be working under the hypothesis in part {\bf (i)}.

\begin{proposition}
\label{conditions Foias wd}
In the setup of the previous definition, the Foias operator is well-defined under any of the following  hypotheses:
\begin{description}
\item [(i)] $ d_X $ is a bounded metric and $g$ is measurable.
\item [(ii)] The input $U$ and state $X$ spaces are compact and $g$ is continuous.
\item [(iii)] The input $U$ space is compact, $g$ is continuous, and the maps $g _{u} $ are all Lipschitz continuous with constants $c _{u} $ such that
$
\sup_{u \in U} \left\{c _{u}\right\}=c< +\infty.
$
\end{description}
\end{proposition}

\noindent\textbf{Proof.\ \ } All that it needs to be shown is that for an element $x _0 \in X $ (and hence for any) and $(\theta, \mu) \in P(U) \times P _1(X)$, the integral
\begin{equation*}
\int _X d_X(x, x _0) dP _g( \theta, \mu) (x)=\int _X \int _U d_X(g _{u}(x), x _0) d \theta(u)d \mu (x)
\end{equation*} 
is finite in the presence of any of the three hypothesis in the statement. It is clearly the case when $ d_X $ is a bounded metric. Under the hypotheses in {\bf (ii)}, the continuous function $d_X(g _{u}(x), x _0) $ reaches a maximum $M>0$ at a point $(u',x') $ and hence
\begin{equation*}
\int _X d_X(x, x _0) dP _g( \theta, \mu) (x)=\int _X \int _U d_X(g _{u}(x), x _0) d \theta(u)d \mu (x) \leq M \int _X \int _U  d \theta(u)d \mu (x) =M< + \infty.
\end{equation*} 
Regarding part {\bf (iii)} we shall show that ${\displaystyle \int _X d_X(x, g_{u _0}(x _0)) dP _g( \theta, \mu) (x)}< + \infty$ for some fixed $u _0 \in U $. By the triangle inequality and the Lipschitz condition
\begin{equation*}
d_X(g_u(x), g_{u _0}(x _0))\leq d_X(g_u(x), g_{u}(x _0))+d_X(g_u(x _0), g_{u_0}(x _0))\leq 
c d_X( x,  x _0)+M,
\end{equation*}
where $M$ is the maximum of the function $d_X(g_u(x _0), g_{u_0}(x _0)) $ thought of as a continuous function of the variable $u $ on the compact set $U$. Consequently,  
\begin{equation*}
{\displaystyle \int _X d_X(x, g_{u _0}(x _0)) dP _g( \theta, \mu) (x)}\leq c \int _Y d_X( x,  x _0) d\mu (x)+M<  + \infty,
\end{equation*}
since $\mu \in P _1(X)$. \quad $\blacksquare$

\section{Driven systems in sequence spaces}
\label{Driven systems in sequence spaces}

In this section we study how, even in the absence of the USP, driven systems induce natural maps between input and output sequence spaces that will be used later on in the paper. 

\noindent {\bf Sequence spaces.} We saw in the previous section how driven systems that satisfy the USP naturally induce input / output systems between the corresponding sequence spaces and that is why it is important to look into their mathematical properties. First of all, given a topological space $Y$, we denote the space of bi-infinite and left semi-infinite countable Cartesian products by
\begin{equation*}
\overline{Y} = \prod_{i=-\infty}^\infty Z_i \quad \mbox{and} \quad \overleftarrow{Y} = \prod_{i=-\infty}^{-1} Z_i, \quad \mbox{respectively, where} \quad Z_i = Y,
\end{equation*}
and equipped with the product topology. Alternatively, we can write $\overleftarrow{Y}= Y^{\Bbb Z ^-}$, where $\Bbb Z ^- $ (respectively $\mathbb{Z}_{-} $)  denotes the set of strictly negative  integer numbers (respectively $\mathbb{Z}^{-}\cup \left\{0\right\} $). Note that there is a natural projection $\pi_{\overleftarrow{Y}}: \overline{Y} \longrightarrow \overleftarrow{Y} $ that extracts from each bi-infinite sequence its left semi-infinite part.
We also note that if $Y$ is a Polish space then so are $\overline{Y} $ and $\overleftarrow{Y} $ since we are considering countable products. Additionally, if $d_Y $ is a metric that makes $Y$ complete and ${\bf w} \in \mathbb{R}^{\mathbb{N}}   $ is a weighting sequence (zero-limit strictly decreasing sequence with $w _1=1 $) then the map $d_{\overleftarrow{Y}}: \overleftarrow{Y} \times \overleftarrow{Y} \rightarrow \overleftarrow{Y} $ given by
\begin{equation}
\label{metric for product}
d_{\overleftarrow{Y}}(\mathbf{x}, {\bf y})=\sup_{i \in \mathbb{N}} \left\{w _i \overline{d_Y}(x _{-i}, y _{-i})\right\}
\end{equation}
induces the product topology on $\overleftarrow{Y} $ and makes it into a Polish space. The symbol $\overline{d_Y} $ denotes the standard bounded metric in $Y$ defined by $\overline{d_Y}( x, y)=\min \left\{d_Y(x,y),1\right\}$. The distance $d_{\overleftarrow{Y}} $ can be used to define a corresponding Wasserstein-1 space $P _1(\overleftarrow{Y}) $ and an associated  Wasserstein distance on it that makes it in turn into a Polish space. This  metric can be easily extended to bi-infinite sequences $\overline{Y} $. Indeed, one first formulates a metric similar to \eqref{metric for product} for the space $\overrightarrow{Y} $ of semi-infinite sequences towards $+ \infty  $ . Then, we write  $\overline{Y} $ as the Cartesian product of $\overleftarrow{Y} $ and $\overrightarrow{Y} $, and we finally put together those two metrics by taking their maximum, which yields a metric $d_{\overline{Y}} $ for $\overline{Y} $.

Regarding notation, elements in sequence spaces will be written in bold and their entries in normal font. Given  $v\in Y$ and ${\bf y} \in \overleftarrow{Y} $,  we define the concatenation map $\sigma_v : \overleftarrow{Y} \to \overleftarrow{Y}$ by $\sigma_v({\bf y})=(\ldots, y_{-2},y_{-1},v)$. The {\bf concatenated sequence} $\sigma_v({\bf y})$ will be sometimes denoted as $ {\bf y}v $. For any $t \in \mathbb{Z}^{-}$, we define the {\bf projection} $\pi_t:\overleftarrow{Y} \rightarrow Y $ by $\pi _t( {\bf y})=y _t $ and the {\bf time delay} operator $T_{-t}:\overleftarrow{Y} \rightarrow\overleftarrow{Y} $ by $T _{-t}({\bf y}) _s= {\bf y}_{s +t} $. It is easy to see that for any $t _1, t _2 \in \mathbb{Z}^{-} $
\begin{equation*}
\pi_{t _1+ t _2}= \pi_{t _1}\circ T_{t _2}=\pi_{t _2}\circ T_{t _1}.
\end{equation*}
These definitions can be extended to the case in which $\overleftarrow{Y}  $ is replaced by $\overline{Y} $,  in which case one can also consider the case $t \in \Bbb Z  $. 

\medskip

\noindent {\bf Pullback attractors,  reachable sets, and generalized filters.} Any driven system $g: U \times X \to X$ defines, for each bi-infinite input $\mathbf{u} \in \overleftarrow{U}$, a nonautonomous dynamical system $\{g(u_n,\cdot)\}_{n\in \mathbb{Z}}$. This system can be  described succinctly using a two-parameter semigroup  (see, for instance, \cite{Kloeden:Rasmussen}) via the map: 
\begin{align} \label{eq_Process}
  \phi_{\mathbf{u}}(n,m,x) = \begin{cases}
        x \quad \quad \quad \quad \quad \quad \quad \text\quad \quad \quad \quad \quad \quad \:\text{ if $n=m$,}
        \\
        g_{u_{n-1}}\circ \cdots \circ g_{u_{m+1}} \circ g_{u_{m}}(x)  \quad \quad \: \:  \text{ if }  m< n,
        \end{cases}
\end{align} 
which is defined for all integers $m\leq n$, and where $g_u(\cdot) = g(u,\cdot)$. The time-invariance of the driven system $g$ makes this semigroup equivariant with respect to  the action of the time delay operator, that is,
\begin{equation}
\label{equivariance property}
  \phi_{T _{-t}(\mathbf{u})}(n,m,x) = \phi_{\mathbf{u}}(n+t,m+t,x), \  \mbox{for all $t \in \Bbb Z$ and $m\leq n$.}
\end{equation}

Note that the set inclusion $\phi_{\mathbf{u}}(m+2,m,X) \subset \phi_{\mathbf{u}}(m+2,m+1,X)$ holds for all $m$ since  $g_{u_{m+1}} \circ g_{u_{m}}(X) \subset g_{u_{m+1}}(X)$.  More generally, using a similar argument, it is easy to see that  $\phi_{\mathbf{u}}(n,m,X)$ is a decreasing sequence of sets as $m$ decreases since $\phi_{\mathbf{u}}(n,m-1,X) \subset \phi_{\mathbf{u}}(n,m,X)$, for any $m\le n$. Hence, if the entire left-infinite input $\{u_m\}_{m<n}$ had influenced the dynamics of the driven system $g$, the system would have evolved at time $n$ to one of the states in the nested intersection
\begin{equation} \label{Seqn_Xn}
X_n(\mathbf{u}) = \bigcap_{m<n} \phi_{\mathbf{u}}(n,m,X).
\end{equation}

Each $X_n(\mathbf{u})$ is the largest {\bf pullback attractor} associated to the input $\mathbf{u} $ (see, for instance,\cite{Kloeden:Rasmussen,Manjunath:Jaeger2}) 
and asymptotically pulls to it the dynamical flows induced by  $g$ if they have begun sufficiently earlier in the sense that 
\begin{equation} \label{Eq_pullback}
	\lim_{j\to \infty}  d\Big( \phi_{\mathbf{u}}(n,n-j,x), X_n(\mathbf{u})\Big) = 0 
	\end{equation}
 for all $n$ and $x\in X$. 

The sets $X_n(\mathbf{u})$ are also related to the outputs of the driven system that are compatible with the input $\mathbf{u} $. Indeed, it can also be shown (see, for instance, \cite{Kloeden:Rasmussen,Manjunath:Jaeger2}) that for any $n \in \Bbb Z$,
\begin{equation}
\label{solution space}
X_n(\mathbf{u}) =\left.\Big \{x \in X\,\right |\, x = x_n \mbox{ where  $\mathbf{x} \in  \overline{X}$  is compatible with input  $\mathbf{u}$} \Big \}.
\end{equation}
In view of this and \eqref{Eq_pullback}, $X_n(\mathbf{u})$, or equivalently the set of all possible components of a solution at the time $n$, attracts any finite-time into the past dynamical evolution described by  $\phi_{\mathbf{u}}(n,n-j,x)$ asymptotically as $j\to \infty$.  Moreover, the relation \eqref{solution space} allows us to characterize the {\bf reachable} or {\bf accessible set}  by the driven system and a set of inputs at a given time $n \in \Bbb Z $. More explicitly, if we set the inputs to be a subset $S$ of the bi-infinite product space $\overline{U} $, we define for any $n \in \Bbb Z$,
\begin{equation}
\label{reachable set}
X _n(S)=\left.\Big \{x \in X\,\right | x = x_n \mbox{ where  $\mathbf{x} \in  \overline{X}$  is compatible with some   $\mathbf{u} \in S\subset \overline{U}$} \Big \}=\bigcup_{\mathbf{u} \in S} X_n(\mathbf{u}).
\end{equation}
The following proposition describes two elementary but important properties of reachable sets.

\begin{proposition}
\label{prop reachable set}
Let $g$ be a driven system and let $X_n(\mathbf{u})  $ and $X _n(S) $ be the reachable sets at $n \in \Bbb Z $ by the input $\mathbf{u} \in \overline{U}$ and by the input set $S\subset \overline{U} $, respectively. Then,
\begin{description}
\item [(i)]  If the state space $X$ is compact and the driven system $g$ is continuous, then $X_n(\mathbf{u})$ is a non-empty compact subset of $X$ for any $\mathbf{u} \in \overline{U}$ and $n \in \Bbb Z $. As a consequence, the driven system $g$ has at least a solution for any input $\mathbf{u} \in \overline{U}$. 
\item [(ii)] If the input space $S \subset \overline{U} $ is {\bf time-invariant}, that is, $T _{-t} (S)=S $ for any $t \in \Bbb Z $, then $X _n(S)=X _m (S) $ for any $n,m \in \Bbb Z $. The time-invariance hypothesis holds when the input space $S$ is the entire product space $\overline{U} $ in which case we write $X_g(\overline{U})= X _n(\overline{U})$, for all $n \in \Bbb Z$. We refer to $X _g(\overline{U}) \subset X$ as the {\bf reachable set} of $g$.
\end{description}
\end{proposition}

\noindent\textbf{Proof.\ \ (i)} It is a consequence of the characterization of compactness using the finite intersection property \cite[Theorem 26.9]{Munkres:topology}. Indeed, since $X$ is compact and $g$ is continuous,  the sets $\phi_{\mathbf{u}}(n,m,X)$, $n , m \in \Bbb Z $, $m < n $, are closed subsets of $X$. Moreover,  using the nesting property $\phi_{\mathbf{u}}(n,m-1,X) \subset \phi_{\mathbf{u}}(n,m,X)$, $n , m \in \Bbb Z $, $m < n $, that we proved right above \eqref{Seqn_Xn}, we can  conclude that for a given $n \in \Bbb Z $, the family $ \left\{\phi_{\mathbf{u}}(n,m,X)\right\}_{m < n }$ is made out of closed sets that satisfy the finite intersection property. The compactness of $X$ implies that  $X_n(\mathbf{u}) = \bigcap_{m<n} \phi_{\mathbf{u}}(n,m,X) $ is non-empty closed (and hence compact) subset of $X$.

\medskip

\noindent {\bf (ii)} Let  $n , m \in \Bbb Z $ and set $t=m-n $. Then, the equivariance property \eqref{equivariance property} implies that 
\begin{multline*}
X _m(S)=X_{n+t}(S)=\bigcup_{\mathbf{u} \in S} X_{n+t}(\mathbf{u})=\bigcup_{\mathbf{u} \in S}  \left(\bigcap_{m'< n+t} \phi_{\mathbf{u}} \left(n+t, m',X\right) \right)\\=
\bigcup_{\mathbf{u} \in S}  \left(\bigcap_{m< n} \phi_{\mathbf{u}} \left(n+t, m+t,X\right) \right)=
\bigcup_{\mathbf{u} \in S}  \left(\bigcap_{m< n} \phi_{T _{-t}(\mathbf{u})} \left(n+t, m+t,X\right) \right)\\=
\bigcup_{\mathbf{u} \in S}  \left(\bigcap_{m< n} \phi_{\mathbf{u}} \left(n+t, m+t,X\right) \right)= X _n(S),
\end{multline*} 
where in the last line we used the invariance property $T _{-t} (S)=S $. \quad $\blacksquare$

\medskip

The concepts that we just introduced allow us to generalize the filter $U _g: \overline{U} \rightarrow \overline{X} $ that we defined in \eqref{filter with USP} and that one can associate to a driven system in the presence of the USP. When that hypothesis does not hold anymore, we can still define a {\bf generalized filter} $U _g: \overline{U} \rightarrow \mathcal{P}(\overline{X}) $ that we  denote with the same symbol and that associates to each input sequence $\mathbf{u} \in \overline{U} $ the set of  solution sequences $U _g(\mathbf{u}) \in \mathcal{P}(\overline{X}) $. The symbol  $\mathcal{P}(\overline{X}) $ denotes the power set. Note that, by definition,
\begin{equation}
\label{charac generalized filter}
U _g(\mathbf{u}) _n=\pi_n(U _g(\mathbf{u}))= X _n( \mathbf{u}), \quad \mbox{for all $n \in \Bbb Z $}.
\end{equation}
This object, introduced in \cite{manjunath:prsl} under the name of {\bf input-representation map},  does generalize the filter introduced in \eqref{filter with USP} since when the USP holds, $U _g(\mathbf{u}) $ is just the singleton containing the unique sequence in $\mathbf{x} \in \overline{X} $ that solves the recursions \eqref{filter with USP} for each given $\mathbf{u} \in \overline{U} $, moreover $U _g(\mathbf{u}) _n=X _n( \mathbf{u})= \{x  _n\} $, for each $n \in \mathbb{Z} $. 

When the state space $X$ is compact, additional properties of the generalized filter $U _g $ (in the sequel we will just call it the filter) can be proved. In that case, as we show in the next proposition, $U _g$ maps $ \overline{U} $ into the set  $\mathcal{F}(\overline{X})\subset  \mathcal{P}(\overline{X}) $  of non-empty compact subsets of $\overline{X} $, which is a metric space on its own using the {\bf Hausdorff distance} $d_H:\mathcal{F}(\overline{X}) \times \mathcal{F}(\overline{X}) \rightarrow  \mathbb{R} $ defined by
\begin{equation*}
d_{\mathrm{H}}(X, Y)=\max \left\{\sup _{x \in X} d_{\overline{X}}(x, Y),\, \sup _{y \in Y} d_{\overline{X}}(X, y)\right\}
\end{equation*}
where $d_{\overline{X}}$ is the metric that we spelled out after \eqref{metric for product}. A general fact about the Hausdorff distance is that since $(\overline{X},d_{\overline{X}}) $ is a compact Polish space, then $(\mathcal{F}(\overline{X}),d_{\mathrm{H}})$ is also compact and complete.

\begin{proposition}
\label{generalized filter x compact}
Let $g:U \times X \rightarrow X $ be a continuous driven system and suppose that the state space $X$ is compact. Then:
\begin{description}
\item [(i)] The corresponding filter $U _g: \overline{U} \rightarrow  \mathcal{P}(\overline{X}) $ is such that $U _g(\overline{U}) \subset \mathcal{F}(\overline{X}) $.
\item [(ii)] Suppose that, additionally, the functions $g _{u}: X \rightarrow X $ do not map any set with uncountably many points contained in $X$ to a single element in $X$, for any $u \in U $. Then, the restricted map $U _g: \overline{U} \rightarrow  (\mathcal{F}(\overline{X}), d_H)$ is continuous if and only if $g$ has the USP for any input in $\overline{U} $.
\end{description}
\end{proposition}

\noindent\textbf{Proof.\ \ (i)} We have to show that for any $\mathbf{u} \in \overline{U} $ the corresponding solution set $U _g(\mathbf{u}) $ is a non-empty and compact subset of $\overline{X}$. Part {\bf (i)} in Proposition \ref{prop reachable set} guarantees that $U _g(\mathbf{u}) $ is non-empty. Regarding the compactness, let $\Gamma_{\mathbf{u}}: \overline{X} \rightarrow \overline{X} $ be the continuous map defined by $(\ldots, x _{-1}, x _0 , x _1, \ldots)\mapsto (\ldots, g_{u_{-2}}(x_{-2}),g_{u_{-1}}(x_{-1}), g_{u_{-0}}(x_{-0}), \ldots) $. It is clear that $U _g(\mathbf{u}) $ coincides with the fixed points set of the continuous map $\Gamma_{\mathbf{u}} $ which since $\overline{X} $ is Hausdorff it is necessarily closed. Since $\overline{X} $ is compact then that fixed point set, and hence $U _g(\mathbf{u}) $, is necessarily compact.
 {\bf (ii)} has been proved in \cite[Theorem 3.1]{manjunath:prsl}. \quad $\blacksquare$ 

\noindent {\bf Driven systems in sequence spaces.} We now introduce  driven systems in sequence spaces induced by the original driven system $g:U \times X \rightarrow X $. These objects will be central in the next developments in the paper.
\begin{definition} 
\label{eq_G}
Let $g:U \times X \to X$ be a driven system. We define its {\bf extension} $\mathbf{G} : \overleftarrow{U} \times \overleftarrow{X} \to \overleftarrow{X}$ {\bf  to sequence space}  by 
\begin{equation}
\label{expression eq_G}
\mathbf{G}(\mathbf{u},\mathbf{x}) = (\ldots,g(u_{-2},x_{-2}), g(u_{-1},x_{-1})).
\end{equation}
\end{definition}

%

Observe first that the semi-infinite solutions of the system associated to a driven system $g:U \times X \to X$ are exactly the fixed points of the map $T _1\circ \mathbf{G}  $ with $T _1 $ the one-lag delay map. More specifically,   $\mathbf{x}\in \overleftarrow{X} $ is a solution for the input $\mathbf{u}\in \overleftarrow{U} $ if and only if
\begin{equation}
\label{fixed points and solutions}
T _1\circ \mathbf{G}(\mathbf{u},\mathbf{x})=\mathbf{x}.
\end{equation} 
In what follows, we focus on the driven system associated to $\mathbf{G} $ in sequence space, its solutions, and their relation with those of the original driven system $g$. Given a sequence (of sequences) $\{{\bf u}_n\}_{n\in \mathbb{Z}}$ with elements in $\overleftarrow{U}  $, we say that the sequence (of sequences)  $\{{\bf x}_n\}_{n\in \mathbb{Z}}$ with elements in $\overleftarrow{X}  $ is a {\bf solution of} $\mathbf{G}$ for the input $\{{\bf u}_n\}_{n\in \mathbb{Z}}$ when 
\begin{equation}
\label{state space in seq}
\mathbf{x}_{n+1}= \mathbf{G}(\mathbf{u}_n, \mathbf{x} _n), \quad \mbox{for all $n \in \Bbb Z $.}
\end{equation} 

In the following paragraphs we discuss the relation between the solutions of the driven systems associated to $g$ and $\mathbf{G} $. We start with the next proposition, which shows that whenever we know that solutions exist for the $g$ and $\mathbf{G} $ systems for all inputs, then the USP of one is equivalent to the USP of the other.

\begin{proposition} 
\label{Prop_solns_existence}
Let $g:U \times X \to X$ be a driven system and let $\mathbf{G} : \overleftarrow{U} \times \overleftarrow{X} \to \overleftarrow{X}$ be its extension to sequence space. 
Suppose that these systems are such that for every input there exists at least one solution. 
Then, $g$ has the USP if and only if $\mathbf{G}$ has the USP.
\end{proposition}

\noindent\textbf{Proof.\ \ } We first prove that if  $g$ has the USP then so does $\mathbf{G}$. Assume that $\mathbf{G}$ does not have the USP.  This means that there exists an input $\{{\bf u}_n\}_{n\in \mathbb{Z}}$ with elements in $\overleftarrow{U}  $ for which there are two distinct solutions $\{{\bf x}_n\}_{n\in \mathbb{Z}}, \{{\bf y}_n\}_{n\in \mathbb{Z}}$ with elements in $\overleftarrow{X}  $. This implies that  So $(\mathbf{x}_m)_k \not= (\mathbf{y}_m)_k$, for some $k \in \mathbb{Z}^-$ and $m\in \mathbb{Z}$. Now, since $\{{\bf x}_n\}_{n\in \mathbb{Z}}$ and $\{{\bf y}_n\}_{n\in \mathbb{Z}}$ are both solutions for the $\mathbf{G}$-system \eqref{state space in seq}, they  satisfy that: $g((\mathbf{u}_n)_j, (\mathbf{x}_n)_j) = (\mathbf{x}_{n+1})_j$ and $g((\mathbf{u}_n)_j, (\mathbf{y}_n)_j) = (\mathbf{y}_{n+1})_j$, for all $j \in \mathbb{Z}^-$ and all $n\in \mathbb{Z}$. In particular, since $(\mathbf{x}_m)_k \not= (\mathbf{y}_m)_k$, this implies that there are two different solutions $\{(\mathbf{x}_n)_k)\}_{n\in \mathbb{Z}}$ and $\{(\mathbf{y}_n)_k)\}_{n\in \mathbb{Z}}$ of the $g$-system for the same input $\{(\mathbf{u}_n)_k)\}_{n\in \mathbb{Z}}$ which contradicts the hypothesis that $g$ has the USP. 

Next, we show that  if $\mathbf{G}$ has the USP then $g$ has the USP. By contradiction, suppose that $g$ does not have the USP and let  let $\mathbf{x}, {\bf y} \in \overline{X}$ be two distinct solutions for the same input $\mathbf{u} \in \overline{U}$. Define the input $\{{\bf u}_n\}_{n\in \mathbb{Z}}$ with elements in $\overleftarrow{U}  $ by  $(\mathbf{u}_n)_j = u_n$, for all $j \in \mathbb{Z}^-$.  Also, define the sequences $\{{\bf x}_n\}_{n\in \mathbb{Z}}, \{{\bf y}_n\}_{n\in \mathbb{Z}}$ with elements in $\overleftarrow{X}  $ by $(\mathbf{x}_n)_j = x_n$ and $(\mathbf{y}_n)_j = y_n$, for all $j \in \mathbb{Z}^-$. Clearly, by definition of $\mathbf{G}$ we have that $\mathbf{G}(\mathbf{u}_n, \mathbf{x}_n) = \mathbf{x}_{n+1}$ and 
$\mathbf{G}(\mathbf{u}_n, \mathbf{y}_n) = \mathbf{y}_{n+1}$, for all $n \in \mathbb{Z}$.  This implies there are two different solutions of $\mathbf{G}$ for the same input, which implies 
$\mathbf{G}$ does not have the USP. \quad $\blacksquare$

\medskip

The existence of solutions hypotheses on $g$ and $\mathbf{G}$ in the previous proposition can be ensured under very general hypotheses. For instance, if the state space $X$ is compact  and convex, it can be shown \cite[Theorem 3.1{\bf (i)}]{RC7} that the $g$ and the $\mathbf{G} $-systems have solutions for any input. This is a consequence of Schauder's Fixed Point Theorem (see \cite[Theorem 7.1, page 75]{Shapiro:Farrago}) when the product topology is used in the corresponding sequence spaces. An extension of this result to a non-compact framework can be found in \cite[Theorem 7{\bf (ii)}]{RC9}. It is worth emphasizing that much like autonomous systems defined on  unbounded spaces exhibit interesting dynamics on bounded invariant sets, the relevant  dynamics of many non-autonomous systems induced by driven systems is contained in bounded absorbing sets \cite{Kloeden:Rasmussen}. This all implies that the hypothesis in the previous proposition on the existence of solutions is in practice not a strong one.

Having said all this, another equivalence result for the equivalence of the USP for $g$ and $\mathbf{G}$ can be formulated in which there is not need to invoke an {\it a priori} knowledge on the existence of solutions for them. The price to pay for this added generality is the restriction the inputs for the $\mathbf{G} $ system to what we call {\bf time-folded} inputs, a notion that we introduce in the next definition.

\begin{definition}
Let $ \left\{\mathbf{x} _n\right\}_{n \in \Bbb Z} $ be a sequence of elements in the space $\overleftarrow{X} $  of left semi-infinite sequences in $X$. We say that the sequence (of sequences) $ \left\{\mathbf{x} _n\right\}_{n \in \Bbb Z} $ is {\bf time-folded} whenever
\begin{equation}
\label{time invariance in seq space}
T _{-t} \mathbf{x} _n = \mathbf{x}_{n+t}, \quad\mbox{for all $t \in \mathbb{Z}_{-} $ and $n \in \Bbb Z $.}
\end{equation}
\end{definition}
The time delay operator $T _{-t}:\overleftarrow{X} \longrightarrow \overleftarrow{X} $, $t \in \mathbb{Z}_{-} $, in the definition is the one that was already introduced at the end of Section \ref{Preliminaries}, in view of which, the time-folding relation \eqref{time invariance in seq space} can be rewritten as
\begin{equation*}
\left(\mathbf{x} _n \right) _{s+t} = (\mathbf{x}_{n+t}) _s, \quad\mbox{for all $t,s \in \mathbb{Z}_{-} $ and $n \in \Bbb Z $.}
\end{equation*}

The following lemma shows that time-folded sequences in $\overleftarrow{X} $ have a very simple structure and that all their terms can be constructed out of a single element in $\overline{X} $.

\begin{lemma}
\label{time folded with vectors}
Let $ \left\{\mathbf{x} _n\right\}_{n \in \Bbb Z} $ be a time-folded sequence of elements in $\overleftarrow{X} $. Then, there exists a unique sequence $\mathbf{x} ^0 \in \overline{X} $ such that 
\begin{equation}
\label{characterization ti}
\mathbf{x} _n= \mathbf{x} ^0_{(- \infty , n]}, \quad \mbox{for all $n \in \Bbb Z$,}
\end{equation}
where $\mathbf{x} ^0_{(- \infty , n]} \in \overleftarrow{X} $ is defined by $\mathbf{x} ^0_{(- \infty , n]} = (\ldots,x^0_{n-1},x^0_{n}) $ or, equivalently, by
\begin{equation}
\label{definition segment}
\left(\mathbf{x} ^0_{(- \infty , n]}\right) _j=x^0_{n+j+1}, \quad \mbox{for all $j \in \mathbb{Z}^{-}$.}
\end{equation}
\end{lemma}

\noindent\textbf{Proof.\ \ } Let $\mathbf{x} ^0 \in \overline{X} $ be the sequence defined by
\begin{equation}
\label{defx0}
x^0_n= \left(\mathbf{x} _n\right)_{-1}, \quad \mbox{for all $n \in \Bbb Z $}.
\end{equation}
We now verify that the invariance condition of $ \left\{\mathbf{x} _n\right\}_{n \in \Bbb Z} $ implies the relation \eqref{characterization ti}. Indeed, for any $j \in \mathbb{Z}^{-} $ and $n \in \Bbb Z  $ we have that 
\begin{equation*}
\left( \mathbf{x} _n\right)_j= \left( T_{-(j+1)} \mathbf{x} _n \right)_{-1} = \left(\mathbf{x}_{n+j+1}\right) _{-1}=x^0_{n+j+1}=\left(\mathbf{x} ^0_{(- \infty , n]}\right) _j,
\end{equation*}
and hence $\mathbf{x} _n= \mathbf{x} ^0_{(- \infty , n]}$,  for all $n \in \Bbb Z$, as required. In the previous expression, the first equality follows from the definition of the time delay operator, the second one follows from the time-folding hypothesis,  the third one from \eqref{defx0}, and the last one from \eqref{definition segment}. \quad $\blacksquare$ 

\begin{proposition} 
\label{Prop_solns}
Let $g:U \times X \to X$ be a driven system and let $\mathbf{G} : \overleftarrow{U} \times \overleftarrow{X} \to \overleftarrow{X}$ be its extension to sequence space. Then $\mathbf{x}=\{x_n\}_{n\in \mathbb{Z}}$ is a solution of $g$ for the input $\mathbf{u}=\{u_n\}_{n\in \mathbb{Z}}$ if and only if the sequence
$\{\mathbf{x}_{(-\infty,n]}\}_{n\in \mathbb{Z}}$ in $\overleftarrow{X} $ is a solution of $\mathbf{G}$ for the input sequence $\{\mathbf{u}_{(-\infty,n]}\}_{n\in \mathbb{Z}}$ in $\overleftarrow{U} $, where 
$\mathbf{x}_{(-\infty,n]} $ and $\mathbf{u}_{(-\infty,n]}$ are defined as in \eqref{definition segment}. Consequently, the driven system $g$ has the USP if and only if the induced map $\mathbf{G}$ in sequence space has the USP when restricted to time-folded inputs.
\end{proposition}

\noindent\textbf{Proof.\ \ }  Suppose first that $\mathbf{x}=\{x_n\}_{n\in \mathbb{Z}}$ is a solution of $g$ for the input $\mathbf{u}=\{u_n\}_{n\in \mathbb{Z}}$, that is, $x_{j+1}=g(u_j,x_j)$ for all $j\in \mathbb{Z}$. We now show that $\{\mathbf{x}_{(-\infty,n]}\}_{n\in \mathbb{Z}}$ is a solution of $\mathbf{G}$ for the input $\{\mathbf{u}_{(-\infty,n]}\}_{n\in \mathbb{Z}}$. To prove this, we just need to verify $\mathbf{G}(\mathbf{u}_{(-\infty,n]},\mathbf{x}_{(-\infty,n]}) =  \mathbf{x}_{(-\infty,n+1])}$ for all $n\in \mathbb{Z}$.  By the definition of  $\mathbf{G}$,  
$\mathbf{G}(\mathbf{u}_{(-\infty,n]},\mathbf{x}_{(-\infty,n]}) = (\ldots,g(u_{n-1},x_{n-1}),g(u_n,x_n))$ for any $n\in \mathbb{Z}$. Since $x_{j+1}=g(u_j,x_j)$ for all $j\in \mathbb{Z}$, we have that  $\mathbf{G}(\mathbf{u}_{(-\infty,n]},\mathbf{x}_{(-\infty,n]}) = (\ldots, x_n,x_{n+1}) = \mathbf{x}_{(-\infty,n+1]}$, as required. 

Conversely, suppose that  $\{\mathbf{x}_{(-\infty,n]}\}_{n\in \mathbb{Z}}$ is a solution of  the extension $\mathbf{G}$, for some input sequence $\{\mathbf{u}_{(-\infty,n]}\}_{n\in \mathbb{Z}}$  in $\overleftarrow{U} $, that is,  $\mathbf{G}(\mathbf{u}_{(-\infty,n]},\mathbf{x}_{(-\infty,n]}) = \mathbf{x}_{(-\infty,n+1]}$, for all $n\in \mathbb{Z}$. By the definition of  $\mathbf{G}$, $\mathbf{G}(\mathbf{u}_{(-\infty,n]},\mathbf{x}_{(-\infty,n]}) = (\ldots,g(u_{n-1},x_{n-1}),g(u_n,x_n))$. Therefore,  $\mathbf{x}_{(-\infty,n+1]} = (\ldots,g(u_{n-1},x_{n-1}),g(u_n,x_n))$, which implies $x_{j+1}=g(u_j,x_j)$, for all $j\le n$. Since  $n$ is arbitrary, we can conclude that the sequence $\mathbf{x}=\{x_n\}_{n\in \mathbb{Z}}$ is a solution of $g$ for the input $\mathbf{u}=\{u_n\}_{n\in \mathbb{Z}}$.  

Finally, since the sequences  $\mathbf{u}=\{u_n\}_{n\in \mathbb{Z}}\in \overline{U}$ and $\mathbf{x}=\{x_n\}_{n\in \mathbb{Z}}\in \overline{X}$ uniquely determine the time-folded sequences $\{\mathbf{u}_{(-\infty,n]}\}_{n\in \mathbb{Z}}$ and $\{\mathbf{x}_{(-\infty,n]}\}_{n\in \mathbb{Z}}$, respectively, and vice-versa, the two implications that we just proved imply that $g$ has the USP if and only if  $\mathbf{G}$ has the USP when restricted to time-folded inputs.  \quad $\blacksquare$

\begin{remark}
\normalfont
If in the last statement in the previous proposition we drop the restriction to time-folded inputs, the claim is in general false (unless we add the existence of solutions property as a hypothesis as we did in Proposition \ref{Prop_solns_existence}). More specifically, even if $g  $ has the USP, the system in sequence space induced by the corresponding $\mathbf{G}  $ may not have that property. As an example, consider the one-dimensional linear system $g(u,x)=ax+u  $, $|a|<1 $, for which $X=U= \mathbb{R}$ and the sequence of (non-time-folded) input sequences given by ${\displaystyle (\mathbf{u} _n) _t= {n ^2}/{t ^2}} $, $n \in \Bbb Z $, $t \in \mathbb{Z}^{-} $. Note that for any $n \in \Bbb Z $, the system induced by $g$ has a unique solution $\mathbf{x} _n \in \overleftarrow{X} $ associated given by ${\displaystyle \left(\mathbf{x} _n\right)_t} = n ^2\sum_{j=0}^{\infty} \frac{a ^j}{(t-j)^2}$, $n \in \Bbb Z $, $t \in \mathbb{Z}^{-} $. On the other hand, the solutions $\left\{\overline{\mathbf{x}}_n\right\}_{n \in \Bbb Z} $ for the $\mathbf{G} $ system in sequence space that have $\left\{\mathbf{u}_n\right\}_{n \in \Bbb Z} $ as input satisfy that $\overline{\mathbf{x}} _{n+1}= \mathbf{G}(\overline{\mathbf{x}} _n, \mathbf{u} _n) $, for all $n \in \Bbb Z $ or, equivalently, that $a \left(\overline{\mathbf{x}} _n\right) _t+ \left(\mathbf{u} _n\right) _t= \left(\overline{\mathbf{x}} _{n+1}\right) _{t} $, for all $n \in \Bbb Z $ and $t \in \mathbb{Z}^{-} $. This relation implies that if a solution $\left\{\overline{\mathbf{x}}_n\right\}_{n \in \Bbb Z} $ exists, it must satisfy that  ${\displaystyle \left(\overline{\mathbf{x}} _n\right)_t} = \frac{1}{t ^2}\sum_{j=0}^{\infty} {a ^j}{(n-j)^2}$. Since this series is divergent, we can conclude that the system induced by $\mathbf{G} $ does not hence have the USP for these inputs.
\end{remark}

Proposition~\ref{Prop_solns} allows us to characterize the reachable set of $\mathbf{G}  $ in terms of the reachable set of $g$. More specifically, the next corollary shows that the reachable set of $\mathbf{G}  $ with time-folded inputs are the time-folded sequences of sequences in $\overleftarrow{X} $ associated to the  solutions of $g$ via the correspondence in Lemma \ref{time folded with vectors}. In order to formulate this precisely we need some notation: given the set $U$, we denote by 
\begin{equation*}
F \left(\overleftarrow{U}\right)= \left\{ \left\{\mathbf{u}_{(- \infty, n]}\right\}_{n \in \Bbb Z}\mid \mathbf{u} \in \overline{U}\right\}
\end{equation*}
the space of time-folded sequences in $\overleftarrow{U}$.

\begin{corollary}
Let $g:U \times X \to X$ be a driven system and let $\mathbf{G} : \overleftarrow{U} \times \overleftarrow{X} \to \overleftarrow{X}$ be its extension to sequence space. Using the notation introduced in \eqref{solution space} and in Propositions \ref{prop reachable set} and \ref{generalized filter x compact}, let $\overleftarrow{X}_{\mathbf{G}}\left(F \left(\overleftarrow{U}\right)\right) $ be the reachable set of the state-space system \eqref{state space in seq}. Then,
\begin{equation}
\label{characterization reachable set G}
\overleftarrow{X}_{\mathbf{G}}\left(F \left(\overleftarrow{U}\right)\right)=\pi_{\overleftarrow{X}}(U_g(\overline{U})),
\end{equation}
where $\pi_{\overleftarrow{X}}: \overline{X} \longrightarrow \overleftarrow{X}$ is the natural projection introduced in Section \ref{Preliminaries}.
\end{corollary}

\noindent\textbf{Proof.\ \ } We note first that the input set $F \left(\overleftarrow{U}\right)  $ is  time-invariant because for any  $\left\{\mathbf{u}_{(- \infty, n]}\right\}_{n \in \Bbb Z} \in F \left(\overleftarrow{U}\right) $ one has that
\begin{equation*}
T _m \left(\left\{\mathbf{u}_{(- \infty, n]}\right\}_{n \in \Bbb Z}\right)=\left\{T _m \left(\mathbf{u}\right)_{(- \infty, n]}\right\}_{n \in \Bbb Z} \in F \left(\overleftarrow{U}\right), \quad \mbox{for any $m \in \Bbb Z $.} 
\end{equation*}
Part {\bf (ii)} of Proposition \ref{prop reachable set} and this time invariance imply that the reachable sets of $ \mathbf{G} $ do not depend on the time step at which they are computed. Taking this into account, we note that for any element $\overline{\mathbf{x}} \in \overleftarrow{X}_{\mathbf{G}}\left(F \left(\overleftarrow{U}\right)\right) \subset \overleftarrow{X} $, by definition there exist $ \left\{\mathbf{u}_{(- \infty, n]}\right\}_{n \in \Bbb Z} \in F \left(\overleftarrow{U}\right) $ and $\left\{\mathbf{x}_{(- \infty, n]}\right\}_{n \in \Bbb Z} \in F \left(\overleftarrow{X}\right) $ such that  
\begin{equation}
\label{sols for G}
\mathbf{G}(\mathbf{u}_{(-\infty,n]},\mathbf{x}_{(-\infty,n]})  = \mathbf{x}_{(-\infty,n+1]}, \quad \mbox{for all $n \in \Bbb Z $,}
\end{equation}
 and $\overline{\mathbf{x}}= \mathbf{x}_{(- \infty,-1]} $. Now, by Proposition \ref{Prop_solns}, the relation \eqref{sols for G} amounts to $\mathbf{x}=\{x_n\}_{n\in \mathbb{Z}}$ being a solution of $g$ for the input $\mathbf{u}=\{u_n\}_{n\in \mathbb{Z}}$, that is $\mathbf{x} \in U _g( \mathbf{u})$ which ensures that $\overline{\mathbf{x}}=\pi_{\overleftarrow{X}}(\mathbf{x}) \in \pi_{\overleftarrow{X}} \left(U _g( \mathbf{u})\right) \subset \pi_{\overleftarrow{X}} \left(U _g(\overline{U})\right) $ and shows the inclusion
\begin{equation*}
\overleftarrow{X}_{\mathbf{G}}\left(F \left(\overleftarrow{U}\right)\right)\subset \pi_{\overleftarrow{X}}(U_g(\overline{U})).
\end{equation*}
The converse inclusion can be obtained by reversing the argument that we just used. \quad $\blacksquare$

\section{Stochastic contractions and invariant measures for driven systems}
\label{Stochastic contractions and invariant measures for driven systems}

We place ourselves in this section in a setup similar to the one in Definition \ref{def:foias} which ensures the existence of a well-defined Foias operator by using, for instance, the hypotheses that we introduced in Proposition \ref{conditions Foias wd}. The main goal in the following pages is proving the {\bf existence} and {\bf uniqueness} of invariant measures for the Foias operators  in both the state and sequence spaces and the {\bf continuity} of their dependences on the input process. The main tool to achieve this is Banach's Fixed Point Theorem which requires two  conditions, namely contractivity and continuity. These two conditions will be treated for the maps of interest in two different subsections. 

We start by introducing  the notion of {\bf stochastic state contraction} which will be at the core of our developments and whose importance is given by the fact that it is, in general, less restrictive than the standard condition evoked to ensure that \eqref{eq:auxEq27} has the unique solution property (see, for instance, \cite{jaeger2001, RC6, RC7}). This condition is satisfied by many parametric models commonly used in time series analysis (see the Examples \ref{VARMAexample} and \ref{GARCHexample} below taken from \cite{RC10}).

\begin{definition}
Let $g:U \times X \rightarrow X $ be a measurable driven system that has as domain the input $(U, d_U)$ and state $(X, d_X)$ spaces that are assumed to be Polish spaces. Given $\theta \in P(U)$ and  $0<c<1$, we say that $g$ is a {\bf $(\theta,c)$-contraction} when
\begin{equation} 
\label{eqn_contraction}
	\int_{U} d_{X}(g_u(x),g_u(y)) \, d\theta(u) \le c\, d_X(x,y)\quad \mbox{holds for all $x,y \in X$.} \quad
\end{equation}
\end{definition}

\begin{remark}[{\bf Stochastic contractivity without the USP}]
\normalfont
\label{stochastic not esp}
Consider
$g: [0,1] \times [0,1]\rightarrow [0,1]$ defined by $g(u,x) = ux$, where subsets of $\mathbb{R}$ are endowed with standard Euclidean metric. The system does not have the USP since for an input sequence comprising of just ones, every constant sequence contained in $[0,1]$ is a solution for that input. On the other hand,  if $\theta = \delta_0$, that is, $\theta$ is a point measure at $0$, we observe that $\int_U d_X(g_u(x),g_u(y)) \, d\theta(u) = 0 < c \, d_X(x,y)$, for every value of $c>0$.
\end{remark}

\begin{example}[{\bf VARMA process with time-varying coefficients}]
\label{VARMAexample}
\normalfont
Suppose ${\bf X}= ({X}_t)_{t \in \Bbb Z}$, with $X _t \in \mathbb{R}^N $, is a vector autoregressive process of first order with time-varying coefficients on $\mathbb{R}^N $ endowed with the Euclidean metric, which we write as:
\begin{equation} 
\label{eq:auxEq27} 
{X}_t = { A}(u_{t-1}) {X}_{t-1}+ f(u_{t-1}), \quad t \in \mathbb{Z},  
\end{equation} 
where $f: \mathbb{R}^d \rightarrow \mathbb{R}^N $ is a measurable map and $\mathbf{u}=(u_t)_{t \in \mathbb{Z}} \sim {\rm IID}$ with  $u_t \in \mathbb{R}^d$, that is, $\mathbf{u}  $ is a ${\Bbb R}^d $-valued sequence of independent and identically distributed random variables. The matrix ${ A}(u) \in \mathbb{M}_{N}$ is assumed to satisfy that ${\rm E}\left[\vertiii{A(u)} \right]< 1 $, where $\vertiii{A(u)} $ denotes the operator norm with respect to the Euclidean metric in $\mathbb{R}^N $ (recall that $\vertiii{A(u)} = \sigma_{{\rm max}}(A)$, the top singular eigenvalue of $A$).

The recursions \eqref{eq:auxEq27} can be encoded as a driven system of the form \eqref{eq_DrS} by defining $g: \mathbb{R}^d \times {\Bbb R}^N \rightarrow {\Bbb R}^N $ as $g(u,x)=A(u)x+ f(u) $. Let now $\theta \in P(U)$ be the law of the components of $\mathbf{u} $. Then, $g$ is a $(\theta,{\rm E}\left[\vertiii{A(u)} \right])$-contraction. Indeed, in this case:
\begin{multline*}
\int_{U} d_{X}(g_u(X),g_u(Y)) \, d\theta(u)=\int_{U} \left\|A(u)(X-Y)\right\| \, d\theta(u)\\
\leq \int_{U} \vertiii{A(u)}\left\|X-Y\right\| \, d\theta(u)
 \le {\rm E}\left[\vertiii{A(u)}\right]\, d_X(X,Y).
\end{multline*}
We emphasize that the condition ${\rm E}\left[\vertiii{A(u)} \right]< 1 $ is in general less restrictive than $\vertiii{A(u)}<1  $,  for all $u \in U $, which would be the standard condition evoked to ensure that \eqref{eq:auxEq27} has the unique solution property (see, for instance, \cite[Theorem 3.1]{RC7}).
\end{example}

\begin{example}[{\bf GARCH process}]
\label{GARCHexample}
\normalfont
We now consider a particular case of the model introduced in \eqref{eq:auxEq27} which is extensively used to describe and eventually to forecast the volatility of financial time series, namely the generalized autoregressive conditional heterostedastic (GARCH) family \cite{engle:arch, bollerslev:garch, Francq2010}.
The GARCH(1,1) model given by the following equations:
\begin{align} 
r_t & = \sigma_t u_{t-1}, \quad u_t \sim {\rm IID}(0, 1), \quad t \in \mathbb{Z}
\label{retgarch}\\ 
\sigma_t^2 & = \omega + \alpha r_{t-1}^2 + \beta \sigma_{t-1}^2, \quad t \in \mathbb{Z} \label{volgarch} \end{align}
with parameters that satisfy $\alpha, \beta, \omega \geq 0$, $\alpha+\beta <1$. 
We now show that the GARCH(1,1) process in \eqref{retgarch}-\eqref{volgarch} falls in the framework  introduced in the previous example. Indeed, let $d=2$ and define 
\[  X_t = \begin{pmatrix}
  r_t^2 \\
   \sigma_t^2    
 \end{pmatrix}, \quad  f(u _t):= \begin{pmatrix}
   \omega u_t^2 \\
    \omega    
  \end{pmatrix}, \quad { A}(u _t)= \begin{pmatrix}
    \alpha u_t^2 & \beta u_t^2 \\
     \alpha & \beta   
   \end{pmatrix},  \quad t \in \mathbb{Z}.
  \]
It is easy to verify that with this choice  one has $ {\rm E}[\vertiii{ A(u )}] = {\rm E}[\alpha u^2 + \beta] = \alpha + \beta < 1$. In this case it is particularly obvious that the condition $ {\rm E}[\vertiii{ A(u )}]  < 1$ is much less restrictive than $\vertiii{ A(u )}=\alpha u^2 + \beta< 1 $,  for all $u \in U $, which is false, for instance when the innovations $u _t $ are not bounded.
\end{example}

\subsection{Stochastic state contractivity and contractive Foias operators}
\label{Stochastic state contractivity and contractive Foias operators}

\begin{center}\begin{figure}[h!]\centering
\includegraphics[scale=0.55]{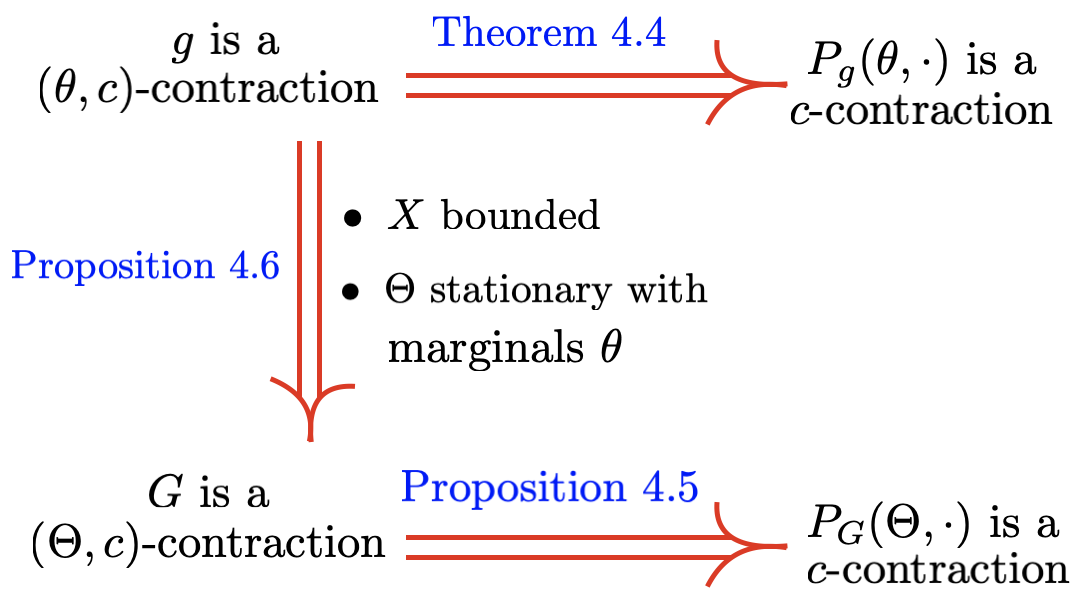}
\caption{Implications between  contractivity in different spaces.} 
\label{Fig_Contraction_Implication}
 \end{figure}
\end{center}

This subsection shows that the stochastic contractivity of a driven system guarantees that its Foias operator is a contraction with respect to the Wasserstein distance (see Theorem \ref{Theorem_contraction}). Moreover, we also spell out conditions, mostly the stationarity of the input process, which guarantee that the corresponding driven system in sequence space and its Foias operator  are also a contraction (see Propositions \ref{prop_stoch_contraction} and \ref{PG is contractive too}). All these different implications are summarized in Figure \ref{Fig_Contraction_Implication}.

\paragraph{Stochastic state contractivity yields contractive Foias operators.} The following results shows that if a driven system is stochastic state contractive with respect to a fixed measure in the input space, then the corresponding Foias operator has the same property.

\begin{theorem} 
\label{Theorem_contraction}
Let $g:U \times X \rightarrow X $ be a measurable driven system that has as domain the input $(U, d_U)$ and state $(X, d_X)$ spaces that are assumed to be Polish. Fix $\theta \in P(U)$  and suppose that  $g$ is a $(\theta,c)$-contraction with $0<c<1$. If $g$ has a well defined Foias operator $P _g: P(U) \times P _1(X) \rightarrow P _1(X) $ then it is necessarily a $c$-contraction with respect to the Wasserstein-1 distance on the second entry, that is,
\begin{equation}
\label{wass-contract}
W(P_g(\theta, \mu_1), P_g(\theta, \mu_2)) \le c \,W(\mu_1,\mu_2), \mbox{ for any $\mu _1, \mu_2 \in P _1(X)$.}
\end{equation}	
\end{theorem}

\noindent\textbf{Proof.\ \ } First of all, given  $f \in \operatorname{Lip} _1(X, \mathbb{R})$, define the function $r_f(x) ={\displaystyle  \frac{1}{c} \int_{U} f(g_u(x)) \, d\theta(u)}$.  It is easy to show that $r_f \in \operatorname{Lip} _1(X, \mathbb{R})$. Indeed, for any $x,y \in X $,
\begin{multline*}
|r_f(x)-r_f(y)| =  \bigg|\frac{1}{c} \int_{U} f(g_u(x)) \, d\theta(u) - \frac{1}{c} \int_{U} f(g_u(y)) \, d\theta(u)\bigg|\\
 \leq \frac{1}{c} \int_{U} |f(g_u(x)) -f(g_u(y))| \, d\theta(u) 
	\leq  \frac{1}{c} \int_{U} d_X(g_u(x) , \, g_u(y)) d\theta(u) \leq d_X(x,y).
\end{multline*}
We now establish \eqref{wass-contract}. Take $\mu _1, \mu_2 \in P _1(X)$ arbitrary. Then:
\begin{multline*}
W(P_g( \theta, \mu_1),P_g( \theta,\mu_2))   =   \sup_{f\in  \operatorname{Lip} _1(X, \mathbb{R})} \bigg(\int_X f(x) d\,P_g(\theta, \mu_1)(x) - \int_X f(x) d\,P_g(\theta, \mu_2)(x)\bigg)
\\
 =  \sup_{f\in  \operatorname{Lip} _1(X, \mathbb{R})} \bigg( \int_X f(x) \, d\Big(\int_{U} \mu_1(g_{u}^{-1}(x)) - \mu_2(g_{u}^{-1}(x)) d \theta(u) \Big)\:\bigg)\\
  =   \sup_{f\in  \operatorname{Lip} _1(X, \mathbb{R})} \bigg(\int_U \int_X f(x) d\Big(\mu_1(g_{u}^{-1}(x)) - \mu_2(g_{u}^{-1}(x))\Big) d\theta(u) \bigg)\\
  =   \sup_{f\in  \operatorname{Lip} _1(X, \mathbb{R})} \bigg(\int_U \int_X f(g_{u}(x)) d\Big(\mu_1(x) - \mu_2(x)\Big) d\theta(u) \bigg), \\
  =   \sup_{f\in  \operatorname{Lip} _1(X, \mathbb{R})} \bigg(
\int_X \Big(\int_U f(g_u(x)) d\theta(u)\Big) d\Big(\mu_1(x) - \mu_2(x)\Big) \bigg)\\
 =  c \sup_{f\in  \operatorname{Lip} _1(X, \mathbb{R})} \bigg(
\int_X r_f(x) d\Big(\mu_1(x) - \mu_2(x)\Big) \bigg)
  \le   c \, W(\mu_1,\mu_2),
\end{multline*} 
where the last inequality follows from the fact that $r_f \in  \operatorname{Lip} _1(X, \mathbb{R}) $. \quad $\blacksquare$

\medskip

The conclusion in the previous theorem can be immediately applied to induced driven systems in sequence spaces, in which case, the metric $d _X  $ in the contractivity condition \eqref{eqn_contraction} has to be replaced by a bounded weighted metric $d_{\overleftarrow{X}} $ of the type that we introduced in \eqref{metric for product}. We shall see later on in Proposition \ref{contract g implies Gk} that the stochastic contractivity in sequence space is naturally inherited under very general hypotheses from a stochastic contractivity hypothesis for the original driven system $g: U \times X \rightarrow X $.

\begin{proposition}
\label{PG is contractive too}
Let $g : U \times X \to X$ be a measurable driven system with Polish input and output spaces and let  ${\bf  G}: \overleftarrow{U} \times \overleftarrow{X} \to \overleftarrow{X}$ be the induced driven system in sequence space  defined in \eqref{eq_G}. The induced system has a well-defined Foias operator $P_{\mathbf{G}}:P( \overleftarrow{U})\times P_1( \overleftarrow{X})\rightarrow P_1( \overleftarrow{X})$ associated with it. Moreover, let $\Theta \in P(\overleftarrow{U})$ and suppose that  ${\bf  G}$ is a $(\Theta,c)$-stochastic contraction, then $P_{{\bf  G}}(\Theta, \cdot )$ is also a $c$-contraction with the Wasserstein-1 metric. 
\end{proposition}

 \noindent\textbf{Proof.\ \ } The operator $P_{\mathbf{G}}:P( \overleftarrow{U})\times P_1( \overleftarrow{X})\rightarrow P_1( \overleftarrow{X})$ is well-defined because of the boundedness of the metric $d_{\overleftarrow{X}} $ in \eqref{metric for product} and part {\bf (i)} of Proposition \ref{conditions Foias wd}. We recall that this metric induces the product topology and makes $\overleftarrow{X} $ into a Polish space. With this in mind, the contractivity claim is a direct corollary of Theorem \ref{Theorem_contraction} that is obtained by replacing $g$ by $\mathbf{G}  $. \quad $\blacksquare$

\medskip

\noindent {\bf Contractive driven systems and their counterparts in sequence spaces.} There are situations in which the previous corollary exhibits a special significance, namely, when the contractivity hypothesis on $\mathbf{G} $ can be obtained out of a contractivity hypothesis on the driven system $g$ that generates it. In the next result we show that is the case when, for instance,  $X$ is bounded and the input process is stationary.

\begin{proposition} 
\label{prop_stoch_contraction}
\label{contract g implies Gk}
Let $g : U \times X \to X$ be a measurable driven system with Polish input and output spaces and let  ${\bf  G}: \overleftarrow{U} \times \overleftarrow{X} \to \overleftarrow{X}$ be the induced driven system in sequence space  defined in \eqref{expression eq_G}. Additionally, suppose that $(X, d_X) $ is bounded and let $\Theta \in P(\overleftarrow{U})$ be the law of a stationary process with marginals $\theta \in P(U)$. Then, if $g $ is a $(\theta, c) $-contraction then $\mathbf{G}   $ is also a $( \Theta, c) $-contraction. More specifically, when in $\overleftarrow{X} $ we consider any weighted metric $d_{\overleftarrow{X}} $ of the type introduced in \eqref{metric for product}, we have that:
\begin{equation} 
\label{eqn_contraction Gk}
	\int_{\overleftarrow{U}} d_{\overleftarrow{X}}(\mathbf{\mathbf{G}} (\mathbf{u},\mathbf{x}),\mathbf{G} (\mathbf{u},{\bf y})) \, d\Theta(\mathbf{u}) \le c\, d_{\overleftarrow{X}}(\mathbf{x},{\bf y}),\  \mbox{ for all $\mathbf{x},{\bf y} \in \overleftarrow{X}$.} 
\end{equation}
\end{proposition}

\noindent\textbf{Proof.\ \ } First of all, the boundedness hypothesis on $X$ allows us to define a weighted metric  \eqref{metric for product} in $\overleftarrow{X} $ without using the  bounded metric $\overline{d_X} $ and by replacing it in the definition just by $d_X $. More explicitly, given the weighted sequence $\mathbf{w} $, the expression  
\begin{equation}
\label{metric for product for compact}
d_{\overleftarrow{X}}(\mathbf{x}, {\bf y})=\sup_{i \in \mathbb{N}} \left\{w _i  {d_X}(x _{-i}, y _{-i})\right\}
\end{equation}
defines a metric on $\overleftarrow{X} $  that induces the product topology. 

Now, since $g$ is a $(\theta, c) $-contraction, we have that for each $t \in \Bbb Z^- $ and each $ \mathbf{x} , {\bf y} \in \overleftarrow{X} $
\begin{equation} 
\label{firs int ineq}
	w _{-t}\int_{U} d_{X}(g_u(x_{t}),g_u(y_{t})) \, d\theta(u) \le c w _{-t}\, d_X(x_{t},y_{t})\leq c\sup_{s \in \Bbb Z^-} \left\{ w _{-s}\, d_X(x_{s},y_{s}) \right\}=c \, d_{\overleftarrow{X}}(\mathbf{x}, {\bf y}).
\end{equation}
Now, in order to prove the claim \eqref{eqn_contraction Gk}, define for fixed $ \mathbf{x}, {\bf y} \in \overleftarrow{X} $ and $t \in \Bbb Z^- $ the sequence of functions
\begin{equation*}
\begin{array}{cccl}
f _t: & \overleftarrow{U} &\longrightarrow & \mathbb{R}\\
	&\mathbf{u} &\longmapsto &\max_{s \in  \left\{t, t+1, \ldots, -1\right\}}\left\{w_{-s}d_X(g_u(x_{s}), g_u(y_{s}))\right\}.
\end{array}
\end{equation*} 
It is clear that 
$$f _t(\mathbf{u})\leq d_{ \overleftarrow{X}}(\mathbf{G}(\mathbf{u},\mathbf{x}), \mathbf{G}(\mathbf{u},\mathbf{y}))   \mbox{ and that } \lim_{t \rightarrow - \infty}f _t(\mathbf{u}) =d_{ \overleftarrow{X}}(\mathbf{G}(\mathbf{u},\mathbf{x}), \mathbf{G}(\mathbf{u},\mathbf{y})).$$
Additionally, $f _{t-1}(\mathbf{u})\geq f _t(\mathbf{u})$, for any $t \in \Bbb Z^- $, and hence the Monotone Convergence Theorem allows us to conclude that
\begin{equation}
\label{with MCT}
\int_{\overleftarrow{U}}d_{ \overleftarrow{X}}(\mathbf{G}(\mathbf{u},\mathbf{x}), \mathbf{G}(\mathbf{u},\mathbf{y}))\, d \Theta(\mathbf{u})
= \lim_{t \rightarrow - \infty}\int_{\overleftarrow{U}}f _t(\mathbf{u})\, d \Theta(\mathbf{u}).
\end{equation}
We now bound ${\displaystyle \int_{\overleftarrow{U}}f _t(\mathbf{u})\, d \Theta(\mathbf{u})}$ by defining, for any $s \in  \left\{t, t+1, \ldots, -1\right\}$, the set $A _s \subset \overleftarrow{U} $ given by
\begin{equation*}
A _s= \left\{ \mathbf{u} \in \overleftarrow{U}\mid f _t(\mathbf{u})=w _{-s}d_X(\mathbf{G}(\mathbf{u},\mathbf{x})_s, \mathbf{G}(\mathbf{u},\mathbf{y})_s)\right\}.
\end{equation*}
In other words, $A _s$ is the measurable subset of $\overleftarrow{U} $ for which the maximum that defines the map $f _t $ is realized for the index $s$. Using this notation and the inequality in \eqref{firs int ineq}, we can write:
\begin{multline*}
 \int_{\overleftarrow{U}}f _t(\mathbf{u})\, d \Theta(\mathbf{u})=\sum_{s=t}^{-1} \int_{A _s}f _t(\mathbf{u})\, d \Theta(\mathbf{u})=
\sum_{s=t}^{-1} \int_{A _s}w_{-s}d_X(\mathbf{G}(\mathbf{u},\mathbf{x})_s, \mathbf{G}(\mathbf{u},\mathbf{y})_s)\, d \Theta(\mathbf{u})\\
=
\sum_{s=t}^{-1} \int_{A _s}w_{-s}d_X(g_u(x_{s}), g_u(y_{s}))\, d \Theta(\mathbf{u})\leq \sum_{s=t}^{-1} \int_{A _s}c d_{\overleftarrow{X}} \left(\mathbf{x} , {\bf y}\right)\, d \Theta(\mathbf{u})\\
=c d_{\overleftarrow{X}} \left(\mathbf{x} , {\bf y}\right)\sum_{s=t}^{-1} \int_{A _s} d \Theta(\mathbf{u})=c d_{\overleftarrow{X}} \left(\mathbf{x} , {\bf y}\right),
\end{multline*}
that is,
\begin{equation*}
 \int_{\overleftarrow{U}}f _t(\mathbf{u})\, d \Theta(\mathbf{u})\leq c d_{\overleftarrow{X}} \left(\mathbf{x} , {\bf y}\right), \quad \mbox{for all $t \in \Bbb Z^- $.}
\end{equation*}
Consequently, by \eqref{with MCT}:
\begin{equation*}
\int_{\overleftarrow{U}}d_{ \overleftarrow{X}}(\mathbf{G}(\mathbf{u},\mathbf{x}), \mathbf{G}(\mathbf{u},\mathbf{y}))\, d \Theta(\mathbf{u})
= \lim_{t \rightarrow - \infty}\int_{\overleftarrow{U}}f _t(\mathbf{u})\, d \Theta(\mathbf{u})\leq c d_{\overleftarrow{X}} \left(\mathbf{x} , {\bf y}\right),
\end{equation*}
as required. \quad $\blacksquare$

\subsection{Continuity of driven systems and their Foias operators}
\label{Continuity of driven systems and their Foias operators}

Given a driven system $g:U \times X \rightarrow X $ between Polish spaces for which the Foias operator $P _g$ exists (see, for instance, the conditions in Proposition \ref{conditions Foias wd}) it is natural to study if the eventual continuity of the driven system induces the same property in $P _g $. More specifically, if we consider the restriction  $P _g: P_1(U) \times P _1(X) \rightarrow P _1(X) $ to input measures in $P_1(U) $, then both the domain and the target of $P _g $ are endowed with the Wasserstein-1 metric and hence the continuity of this map can be studied. The following uniform continuity hypothesis will be needed in the sequel.

\begin{definition}
Let $g:U \times X \rightarrow X $ be a driven system with  Polish input and output spaces. We say that $g$ is {\bf uniformly continuous on the first entry} if for any $\epsilon > 0  $ there exists $\delta(\epsilon)>0 $ such that  if $d_U(u,v)< \delta (\epsilon)$ then  $d_X(g(u,x),g(v,x))< \epsilon $, for all $x \in X $. This definition is extended to the {\bf uniform continuity on the second entry} in a straightforward manner.
\end{definition}

\begin{remark}
\label{unif sim and sep}
\normalfont
Note that if the product $U \times X $ is endowed with the product metric $$d_{U \times X}((u,x),(v,y))=\max  \left\{d _U(u,v), d_X(x,y)\right\}$$ then $g:U \times X \rightarrow X $ is uniformly continuous if and only if it is uniformly continuous  simultaneously on the first and the second entries. 

Indeed, if $g$ is uniformly continuous  for any $\epsilon>0 $ there exists a $\delta_{U \times X}(\epsilon)>0 $ such that 
\begin{equation}
\label{implication of unif cont}
d _X(g(u,x), g(v,y))< \epsilon  \mbox{\  whenever\ } d_{U \times X}((u,x),(v,y))<\delta_{U \times X}(\epsilon). 
\end{equation}
Now, if $x,y \in X$ are such that $d_X(x,y)<\delta_{U \times X}(\epsilon) $ this implies that
$$d_{U \times X}((u,x),(u,y))=d_X(x,y)<\delta_{U \times X}(\epsilon), \quad \mbox{ for any $u \in U $,  }$$
and hence $d _X(g(u,x), g(u,y))< \epsilon $, which proves that $g$ is uniformly continuous on the second entry. It can be analogously shown that it is also uniformly continuous on the first entry. Conversely, the uniform continuity on the first and the second entries implies the uniform continuity of $g$ because for any $u,v \in U $ and $x,y \in X $, the triangle inequality guarantees that
\begin{equation*}
d _X(g(u,x), g(v,y))\leq d _X(g(u,x), g(u,y))+ d _X(g(u,y), g(v,y)).
\end{equation*}
\end{remark}

\begin{theorem} 
\label{Theorem_continuity}
Let $g: U \times X \to X$ be a  measurable driven system with  Polish input and output spaces and $(X, d _X)$ compact. If $g$ is uniformly continuous on the first entry, then the corresponding Foias operator $P _g:P _1(U)\times P _1(X)\rightarrow P _1(X)$ is continuous on the first entry, that is, the maps $P _g(\cdot , \mu):P _1(U)\rightarrow P _1(X)$ are all continuous for any $\mu\in P _1(X)$.
\end{theorem}

Before we start the proof we introduce the following Lemma.

\begin{lemma}
\label{lemma with sf}
Let $g: U \times X \to X$ be a measurable driven system  that is uniformly continuous on the first entry, with  Polish input and output spaces, and $(X, d _X)$ compact. Let $f:X \rightarrow \mathbb{R} $ be a continuous (and hence uniformly continuous) function. Given $\mu \in P(X) $, the map $s _f:U \rightarrow \mathbb{R}  $ defined by
\begin{equation}
\label{definition sf}
s _f (u)=\int _X f(g(u,x)) d\mu(x)
\end{equation}
is continuous and bounded.
\end{lemma}

\noindent\textbf{Proof of the Lemma.\ \ } By the compactness of $X$ and the continuity of $f$, there exists $M>0  $ such that $|f(x)|\leq M $ for all $x \in M $. This implies that for any $u \in U $:
\begin{equation*}
|s _f(u)|= \left|\int _X f(g(u,x)) d\mu(x) \right|\leq  \int _X M d\mu(x)  =M,
\end{equation*}
which proves that $s _f  $ is bounded. We now prove that $s _f  $ is continuous. Let $\epsilon>0 $ and let $\delta( \epsilon)>0  $ be the scalar that by the uniform continuity of $f$ implies that 
\begin{equation}
\label{f cont}
|f(x)-f(y)|< \epsilon \mbox{ whenever $d_X(x,y)< \delta (\epsilon)$}. 
\end{equation}
Let now $\delta'(\epsilon)>0 $ be the element that by the uniform continuity of $g$ on the first entry guarantees that 
\begin{equation}
\label{unif cont g}
d_X(g(u,x),g(v,x))< \delta(\epsilon) \mbox{ for any $x \in X $ and whenever $d_U(u,v)< \delta'(\epsilon)$.} 
\end{equation}
Now, if $u,v \in U $  are such that $d_U(u,v)< \delta'(\epsilon)$ then:
\begin{equation*}
|s _f (u)-s _f (v)|=\left|\int _X \left(f(g(u,x)) - f(g(v,x))\right) d\mu(x)  \right|\leq \int _X \left|\left(f(g(u,x)) - f(g(v,x))\right) \right| d\mu(x) < \epsilon,
\end{equation*}
where the last inequality is a direct consequence of \eqref{unif cont g} and \eqref{f cont}, which proves the continuity of $s _f $. $\blacktriangledown $

\medskip

{\bf Proof of the Theorem.\ \  }We will proceed by using the fact that the Wasserstein distance \eqref{characterization Wasserstein} metrizes  the weak convergence as characterized in \eqref{characterization weak conv}. 
Let $\mu \in  P _1(X) $ be arbitrary but fixed and let $\left\{\theta_n\right\}_{n \in \mathbb{N}}$ be a convergent sequence in $P _1(U) $, that is, there exists $\theta\in P _1(U)$ such that 
\begin{equation}
\label{limit theta}
\lim_{n \rightarrow \infty} W(\theta_n, \theta)=0.
\end{equation}
The continuity property that we are interested in is guaranteed if ${\displaystyle \lim_{n \rightarrow \infty} W(P _g(\theta_n, \mu), P _g(\theta, \mu))=0}$, which by \eqref{characterization weak conv} is established if for any continuous function $f: X \rightarrow \mathbb{R}  $ that satisfies that $|f(x)|\leq C(1+ d_X(x, x _0)) $, we have that
\begin{equation}
\label{limit with pg}
\lim _{n \rightarrow \infty}\int _X f(x)\, dP _g (\theta_n , \mu)(x)=\int _X f(x)\, dP _g (\theta , \mu)(x).
\end{equation}
Using the notation introduced in Lemma \ref{lemma with sf} we rewrite
\begin{multline*}
\int _X f(x)\, dP _g (\theta_n , \mu)(x)=\int _X\int _U f(g(u,x))\, d \theta_n(u)\, d\mu(x)\\=
\int _U\int _X f(g(u,x))\, d\mu(x)\, d \theta_n(u)=\int _U s _f (u)\, d \theta_n(u).
\end{multline*}
Consequently, the equality \eqref{limit with pg} holds whenever
\begin{equation*}
\lim _{n \rightarrow \infty}\int _U s _f (u)\, d \theta_n(u)=\int _U s _f (u)\, d \theta(u),
\end{equation*}
which is the case because by \eqref{limit theta} and by \eqref{characterization weak conv} we have that
\begin{equation*}
\lim _{n \rightarrow \infty}\int _U h (u)\, d \theta_n(u)=\int _U h (u)\, d \theta(u),
\end{equation*}
for all $h:U \rightarrow \mathbb{R}  $ continuous such that $|h(x)|<C(1+ d _U(u _0,u)) $, for some $C>0 $. The map $s_f$ has that property because due to Lemma \ref{lemma with sf} it is continuous and bounded by some constant $M>0$, and hence $|s_f(x)|<M\leq M(1+ d _U(u _0,u)) $. \quad $\blacksquare$

\medskip

The following result extends the continuity statement in the previous theorem to the induced Foias operator $P_{\mathbf{G} }:P( \overleftarrow{U})\times P_1( \overleftarrow{X})\rightarrow P_1( \overleftarrow{X})$ on the sequence space.

\begin{corollary} 
\label{corr_continuity}
Let $g : U \times X \to X$ be a measurable driven system with Polish input and output spaces, and $X$ compact. Let  ${\bf  G} :  \overleftarrow{U} \times  \overleftarrow{X}\rightarrow   \overleftarrow{X}$ be the  induced driven system in sequence space as defined in \eqref{eq_G} 
and assume that  that ${\bf  G} $ is uniformly continuous on the first entry. Then,  the corresponding Foias operator $P_{\mathbf{G} }:P _1 ( \overleftarrow{U})\times P_1( \overleftarrow{X})\rightarrow P_1( \overleftarrow{X})$ is continuous on the first entry.
\end{corollary}

\noindent\textbf{Proof.\ \ } It can be obtained in a straightforward manner from Theorem \ref{Theorem_continuity} by replacing in its statement the driven system $g$ by ${\bf  G} $, which inherits measurability from $g$. Note that if $U$ and $X$  are Polish then so are $\overleftarrow{U}$ and $\overleftarrow{X}$ with the product topology. Moreover, $\overleftarrow{X}$ is also compact because of Tychonoff's Theorem \cite[Theorem 37.3]{Munkres:topology}. Finally, the induced Foias operator $P_{\mathbf{G} }:P( \overleftarrow{U})\times P_1( \overleftarrow{X})\rightarrow P_1( \overleftarrow{X})$ is guaranteed to be well-defined by part {\bf (i)} in Proposition \ref{conditions Foias wd} and by the boundedness of the metric \eqref{metric for product} on the product space. \quad $\blacksquare$

\medskip

In order to apply Corollary~\ref{corr_continuity} on $\mathbf{G}$, we shall now establish conditions on $g$ in the next corollary which guarantee the uniform continuity of $\mathbf{G}$ on the first entry.  Before we state it, we define some general properties that a metric generating the product topology can possess.

\begin{definition}
\label{uniform-product-factor metric}
Let $(Y,d_Y)$ be a metric space and let $d_{\overleftarrow{Y}}$ be a metric that generates the product topology.  
\begin{description}
\item  [(i)]  We say that $d_{\overleftarrow{Y}}$ is a {\bf uniform-product metric} if for any  $\epsilon>0$ there exists a $\delta>0$  such that 
$d_{\overleftarrow{Y}}(\mathbf{y}, \mathbf{z}) < \epsilon$ whenever
$d_Y(y_{-i},z_{-i})< \delta$, for all $i\le -1$ and for all $\mathbf{y}, \mathbf{z} \in \overleftarrow{Y}$. 
\item [(ii)]   We say that $d_{\overleftarrow{Y}}$ is a {\bf uniform-factor metric} if for any  $\beta>0$ there exists an $\alpha>0$ such that  $d_Y(y_{-i},z_{-i})< \beta$ whenever $d_{\overleftarrow{Y}}(\mathbf{y}, \mathbf{z}) < \alpha$, for all $i\le -1$ and for all $\mathbf{y}, \mathbf{z} \in \overleftarrow{Y}$.
\end{description}
\end{definition}

It can be readily verified  that any  metric of the form \eqref{metric for product} is both a uniform-factor and a uniform-product metric. 

\begin{corollary} 
\label{Corr_UC} 
Let $g : U \times X \to X$ be a  driven system with Polish input and output spaces.
Suppose that $d_{\overleftarrow{U}}$ is a uniform-factor metric, that
$d_{\overleftarrow{X}}$ is simultaneously a uniform-product and a uniform-product metric, and that
$g$ is uniformly continuous on the first entry. Then the extension ${\bf  G}$ of $g$  to sequence space defined in \eqref{expression eq_G} is also uniformly continuous on the first entry. 
Additionally, if $g$ is  uniformly continuous, then 
${\bf  G}$ is also uniformly continuous.
\end{corollary}

\noindent\textbf{Proof.\ \ } We first show  that ${\bf  G}$ is uniformly continuous when $g$ is uniformly continuous.    Then the proof of the uniform continuity of ${\bf  G}$  on the first entry follows from the uniform continuity of $g$ on the first entry easily. We proceed in three steps:

{\bf Step 1.} Fix $\epsilon> 0$. Since $d_{\overleftarrow{X}}$  is a uniform-product metric, we can find a $\tau>0$ such that if 
$d_{X}({\bf  G}(\mathbf{u},\mathbf{x})_i, {\bf  G}(\mathbf{v},\mathbf{y})_i) < \tau$ for all $i\le -1$, then 
 $d_{\overleftarrow{X}}({\bf  G}(\mathbf{u},\mathbf{x}), {\bf  G} (\mathbf{v},\mathbf{y})) < \epsilon$. 
Fix any such $\tau>0$.

{\bf Step 2. } When $g$ is uniformly continuous, we can find a $\gamma>0$  independent of $u_i,v_i,x_i,y_i$ and independent of $i \le -1$, so that $d_X(g(u_i,x_i),g(v_i,y_i))< \tau$  whenever $d_U(u_i, v_i) < \gamma$ and $d _X(x _i, y _i)< \gamma  $ (see the implication in \eqref{implication of unif cont}). 

{\bf Step 3. } Since $d_{\overleftarrow{U}}$ and $d_{\overleftarrow{X}}$ are uniform-factor metrics, given $\gamma$ there exits a $\delta>0$ so that if $d_{\overleftarrow{U}}(\mathbf{u},\mathbf{v})< \delta$ and $d_{\overleftarrow{X}}(\mathbf{x},\mathbf{y})< \delta$ then $d_U(u_i, v_i) < \gamma$ and $d_X(x_i, y_i) < \gamma$, for all $i \le -1$. This implies in particular that $d_X(x_i, y_i) < \gamma$, for all $i \le -1 $, whenever $d_{\overleftarrow{U} \times \overleftarrow{X}}(( \mathbf{u}, \mathbf{x}),( \mathbf{v},\mathbf{y}))< \delta$.

Hence, we have from the above three steps that if $d_{\overleftarrow{U} \times \overleftarrow{X}}(( \mathbf{u}, \mathbf{x}),( \mathbf{v},\mathbf{y}))< \delta$,  we then necessarily have that
$d_{\overleftarrow{X}}(\mathbf{G}(\mathbf{u},\mathbf{x}), \mathbf{G}(\mathbf{v},\mathbf{y}) < \epsilon$. 

In particular, when $g$ is only uniformly continuous on the first entry, we can set $\mathbf{x} =  \mathbf{y}$ in the  steps above to obtain the implication $d_{\overleftarrow{U}}(\mathbf{u},\mathbf{v})< \delta$  $\Longrightarrow$ 
$d_{\overleftarrow{X}}(\mathbf{G}(\mathbf{u},\mathbf{x}), \mathbf{G}(\mathbf{v},\mathbf{x}) < \epsilon$.  $\blacksquare$

\medskip

\begin{center}\begin{figure}[h!]\centering
\includegraphics[scale=0.6]{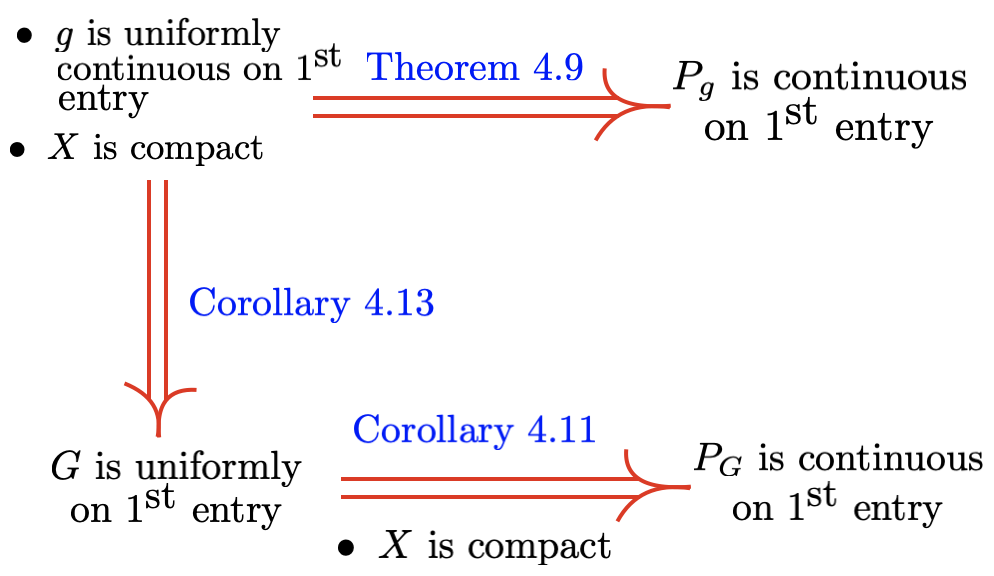}
\caption{Implications of continuity between the different maps.} 
\label{Fig_Continuity_Implication}
 \end{figure}
\end{center}

The implications about the continuity of the different maps that we have proved in this subsection is summarized in Figure \ref{Fig_Continuity_Implication}.

\subsection{The fixed points of the Foias operator}
\label{The fixed points of the Foias operator}

The importance of the contractivity and continuity results in the previous two subsections lies in the fact that they can be used in conjunction with Banach's Fixed Point Theorem to prove, for each input process, the existence of a unique fixed point of the corresponding Foias operator, as well as its continuous dependence on the input process.  The following theorem provides a specific statement in this direction for a driven system $g$ and its Foias operator $P _g  $. We generalize later on this result in Theorem \ref{continuous foias fixed point seq} to the driven system $\mathbf{G} $ in sequence space its Foias operator $P _{\mathbf{G}}  $.

\begin{theorem}[{\bf Fixed points of $P _g$}] 
\label{Theorem_im_continuity}
Consider a measurable driven system $g: U \times X \to X$ with Polish input and output spaces and  $(X, d_X) $ compact. Assume that  for each $\theta \in P(U)$ the map $g$ is a $(\theta, c _\theta)$-stochastic contraction and that one of the following hypotheses holds true:
\begin{description}
\item [(i)] There exists a constant $c _0 \in (0,1)$ such that $0< c_\theta < c_0 < 1$ for all $\theta \in P _1(U)$  and $g$ is uniformly continuous on the first entry.
\item [(ii)] $g$ is uniformly continuous.
\end{description}
Then, for any $\theta \in P_1(U)$ there exists a unique $\mu _\theta\in P _1(X)$ which is a fixed point of the  map $P _g(\theta,\cdot): P _1(X) \to P _1(X)$ and, moreover, the map $S _g:P _1(U)\to P _1(X) $ that assigns $\theta \mapsto \mu_\theta$, is continuous.
\end{theorem}

\begin{center}
\begin{figure}[h!]\centering
\includegraphics[scale=0.5]{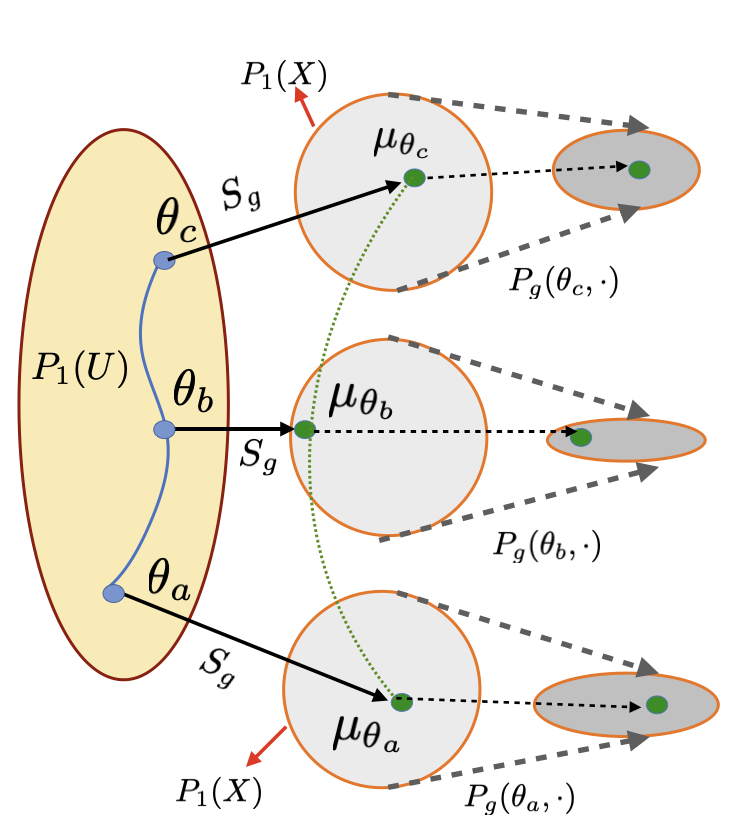}
\caption{Graphical illustration of the map $S _g $,  its continuity, and of its relation with the fixed points of the Foias operator.} 
 \end{figure}
\end{center}

\noindent\textbf{Proof of the theorem.\ \ } Notice first that part {\bf (i)} of Proposition \ref{conditions Foias wd} implies, together with the compactness of $X$ that the Foias map $P _g:P(U) \times  P _1(X) \to P _1(X)$ is well-defined. Fix now $\theta \in P (U)$ and recall that by Theorem \ref{Theorem_contraction}, the $(\theta, c _\theta)$-stochastic contractivity hypothesis on $g$ implies that $P _g(\theta,\cdot): P _1(X) \to P _1(X)$ is a $c _\theta $-contraction with respect to the Wasserstein distance. Since the completeness of $(X, d_X)$ implies that of $(P _1(X),W)$ by  \cite[Theorem 6.18]{villani2009optimal}, Banach's Fixed Point Theorem implies the existence of a unique $\mu_\theta \in P _1(X)$ such that $P _g(\theta,\mu_\theta)= \mu_\theta $ for each $\theta \in P (U)$, as well as the existence of the map $S _g:P (U)\to P _1(X) $ that assigns $\theta \mapsto \mu_\theta$.

The remainder of the proof is dedicated to showing that the restriction $S _g:P_1 (U)\to P _1(X) $ is continuous in the presence of the hypotheses in {\bf (i)} or {\bf (ii)}. 

Assume first that {\bf (i)} holds and suppose that we have a sequence $\left\{\theta_n\right\}_{n \in \mathbb{N}}\subset P _1(U)$ such that  ${\displaystyle \lim_{n \rightarrow \infty}\theta_n= \theta \in P _1(U)}$. It suffices to show that ${\displaystyle \lim_{n \rightarrow \infty} \mu_{\theta_{n}}= \mu_\theta}$.  We now set, for any $n \in \mathbb{N} $,
\begin{equation} 
\label{eqn_cn}
c_n= \inf \left.\bigg\{b\in (0,1) \right| \int_U d_X(g_u(x), g_u(y)) d\theta_n(u) \le b \,d_X(x,y), \: \forall x,y \in X \bigg \}.
\end{equation}
By Theorem~\ref{Theorem_contraction}, we have
\begin{equation} \label{eqn_cont}
	W(P_g(\theta_n, \mu_\theta), P_{g} (\theta_n,\mu_{\theta_{n}})) \le c_n W (\mu_\theta, \mu_{\theta_{n}}).
\end{equation}
We hence have that
\begin{eqnarray}
	W(\mu_\theta,\mu_{\theta_{n}}) & \le & W(\mu_\theta, P_g(\theta_n, \mu_\theta) ) + W(P_g(\theta_n, \mu_\theta), \mu_{\theta_{n}}) \hspace{1cm} \mbox{(by the triangle  inequality)} \nonumber \\
	& = &  W(\mu_\theta, P_g(\theta_n, \mu_\theta) )  + W(P_g(\theta_n, \mu_\theta), P_{g} (\theta_n,\mu_{\theta_{n}})) \hspace{0.5cm} \mbox{(since  $\mu_{\theta_{n}}= P_{g} (\theta_n,\mu_{\theta_{n}})$)}\nonumber \\
	& \le & \frac{1}{1-c_n} W(\mu_\theta, P_g(\theta_n, \mu_\theta) ) \hspace{1cm} \mbox{(by \eqref{eqn_cont}).}  \label{eq_ineq} 
\end{eqnarray}
Notice now that the hypotheses in point {\bf (i)} and Theorem \ref{Theorem_continuity} guarantee that the maps $P _g(\cdot , \mu):P _1(U)\rightarrow P _1(X)$ are all continuous, for any $\mu\in P _1(U)$, and hence as $n\to \infty$,  $W(\mu_\theta, P_g(\theta_n, \mu_\theta)) \to 
W(\mu_\theta, P_g(\theta, \mu_\theta)) = W(\mu_\theta,\mu_\theta) = 0$.   In addition,  since ${\displaystyle \limsup_{n \rightarrow \infty} c_n \le c_0 < 1}$, then $\frac{1}{1-c_{n}}$ is bounded above, and hence from  \eqref{eq_ineq}, ${\displaystyle\lim _{n \rightarrow  \infty} W(\mu_\theta,\mu_{\theta_{n}}) = 0}$, as required. 

We next consider the case {\bf (ii)} in which we cannot assume that the contraction constants $c_n$ in \eqref{eqn_cn} are bounded by a $c_0<1$, but  require in exchange that $g$ is uniformly continuous. Notice that if under this hypothesis we have that $\limsup_{n \rightarrow \infty} c_n = c _0<1$, since by Remark \ref{unif sim and sep} the uniform continuity of $g$ implies its uniform continuity on the first entry, this case reduces to the one in part {\bf (i)}. Suppose hence that $\limsup_{n \rightarrow \infty} c_n = 1$. Now, since $g$ is uniformly continuous (and hence continuous) and $X$ is compact, then for some fixed $x,y \in X $,  the map $h_{x,y}:U \rightarrow \mathbb{R} $ given by $h_{x,y}(u)= d_X(g_u(x), g_u(y)) $ is bounded by some constant $M>0$ and hence it satisfies that $|h_{x,y}(u)|<M\leq M(1+ d_U(u, u _0)) $. Given that the sequence $\left\{\theta_n\right\}_{n \in \mathbb{N}}\subset P _1(U)$ is such that  ${\displaystyle \lim_{n \rightarrow \infty}\theta_n= \theta \in P _1(U)}$, the characterization of the weak convergence \eqref{characterization weak conv} implies that
\begin{equation} 
\label{eq_pw}
\lim_{n \rightarrow \infty}	\int_U d_X(g_u(x), g_u(y)) \, d\theta_n(u) = \int_U d_X(g_u(x), g_u(y)) \, d\theta(u) \quad \mbox{for all} \quad x,y \in X.
\end{equation}
Consider now the sequence of functions $f _n:X \times X \rightarrow \mathbb{R} $, $n\in \mathbb{N}$, defined by $f_n(x,y) := \int_U  d_X(g_u(x), g_u(y)) \, d\theta_n(u)$. We shall now show that 
$\{f_n\}_{n \in \mathbb{N}}$ is equicontinuous when in $X \times  X $ we consider the product metric
$d_{X \times X}((x,y),(a,b))=\max  \left\{d _X(x,a), d_X(y,b)\right\}$. An important element in our arguments will be the uniform continuity of $g$ when in its domain $U \times X $ we consider the product metric
$d_{U \times X}((u,x),(v,y))=\max  \left\{d _U(u,v), d_X(x,y)\right\}$. Notice first that for any $(x,y), (a,b)\in X \times X $:
\begin{eqnarray}
|f_n(a,b) - f_n(x,y)|  &=&   
\left|\int_U (d_X(g_u(a), g_u(b)) - d_X(g_u(x), g_u(y)) \, d\theta_n(u) \right| \nonumber \\
&\le&  \int_U \mid (d_X(g_u(a), g_u(b)) - d_X(g_u(x), g_u(y)) \mid \, d\theta_n(u).  \label{eqn_equicont}
\end{eqnarray}
The compactness of $X$ implies that the map $d_{X}:X \times X \to \mathbb{R} $ is uniformly continuous and hence for any $\epsilon> 0 $ there exits $\delta_{d_{X }}(\epsilon)> 0$ such that for any  $(x,y), (a,b)\in X \times X $ such that $d_{X \times X}((x,y), (a,b))<\delta_{d_{X \times X}}(\epsilon)$ we have that $|d_X(x,y)- d_X(a,b)|< \epsilon  $. Additionally, as we saw in Remark \ref{unif sim and sep}, the uniform continuity of $g $ implies its uniform continuity on the second entry and hence for any $\epsilon> 0 $ there exits $\delta_{g}(\epsilon)> 0$ such that $d_X(g_u(x), g_u(y))< \epsilon $ for any $u \in U $ and any $x, y \in X $ such that $d _X(x,y)<\delta_{g}(\epsilon) $. 

We shall now prove the equicontinuity of $\{f_n\}_{n \in \mathbb{N}}$ by showing that for any $\epsilon> 0 $, then $|f_n(a,b) - f_n(x,y)|< \epsilon$, whenever  $d_{X \times X}((x,y), (a,b))<\delta  _g(\delta_{d_{X }}(\epsilon))$. Indeed, by the definition of the product metric $d_{X \times X} $, if  $d_{X \times X}((x,y), (a,b))<\delta  _g(\delta_{d_{X }}(\epsilon))$ then $d_X(x,a)<\delta  _g(\delta_{d_{X }}(\epsilon))$ and $d_X(y,b)<\delta  _g(\delta_{d_{X }}(\epsilon))$  and hence $d_X(g_u(x), g_u(a))< \delta_{d_{X }}(\epsilon)$ and $d_X(g_u(y), g_u(b))< \delta_{d_{X }}(\epsilon)$, for any $u \in U $. This implies that $d_{X \times X}((g_u(x), g_u(y)), (g_u(a), g_u(b)))<\delta_{d_{X }}(\epsilon) $ and hence that 
\begin{equation*}
| d_X(g_u(a), g_u(b)) - d_X(g_u(x), g_u(y)) |< \epsilon \quad \mbox{for all} \quad u \in U.
\end{equation*}
Since $\theta_n$ is a probability measure, from \eqref{eqn_equicont}, we have  that $|f_n(a,b) - f_n(x,y)|< \epsilon$, which proves the equicontinuity of $\{f_n\}$.  

Let now $\{c_{n_{j}}\}$  be a subsequence of $\{c_n\}$ such that $\lim_{j\to \infty} c_{n_{j}} = 1$. By \eqref{eqn_cn} and the compactness of $X$, we can find  a sequence $\{(x_{n_{j}}, y_{n_{j}})\}$ in $X \times X$ such that $f_{n_{j}}(x_{n_{j}}, y_{n_{j}}) = c_{n_{j}}d_X(x_{n_{j}}, y_{n_{j}})$ for all $j\in \mathbb{N}$. Again, since $X$ (and hence $X \times X$) is compact, we assume without loss of generality that $\{(x_{n_{j}}, y_{n_{j}})\}$ converges to some point $(x_0,y_0) \in X \times X$. 

Next, we note that $\{f_{n_{j}}\}$ converges point-wise to $f$ by \eqref{eq_pw}. Also, since $\{f_n\}$ or any of its subsequences are equicontinuous, there is a subsequence that converges uniformly to $f$ by the Arzela-Ascoli theorem, which guarantees the continuity of $f$. Without loss of generality assume that $\{f_{n_{j}}\}$ itself converges uniformly to $f$. Therefore, by the continuity of $f$:
\begin{equation*}
f(x_0,y_0) = 	\lim_{j \to \infty} f_{n_{j}}(x_{n_{j}}, y_{n_{j}})  
 =  	\lim_{j \to \infty} c_{n_{j}}d_X(x_{n_{j}}, y_{n_{j}}) 
 =  d_X(x_0,y_0)
\end{equation*}
which contradicts that $\int_U d(g_u(x_0), g_u(y_0)) \, d\theta(u) < c_\theta d_X(x_0,y_0)$ for some $0<c_\theta<1$.  Hence, $\limsup_{j \rightarrow \infty} c_{n_{j}} < 1$ necessarily and the theorem is proven.    \quad $\blacksquare$

\medskip

\medskip

The goal of our last theorem is showing that the conclusions about the existence of fixed points of the maps $P _g(\theta, \cdot ) $  and their continuous dependence on $\theta$ that we proved in Theorem \ref{Theorem_im_continuity} can be extended to $P _{\mathbf{G}}(\Theta, \cdot ) $ by using hypotheses that are exclusively formulated in terms of $g$, provided that the inputs $\Theta  $ are stationary. We hence denote as
\begin{equation*}
P _S(\overleftarrow{U})= \left\{ \Theta \in P _1(\overleftarrow{U})\mid T _{1\, \ast} \Theta= \Theta\right\}
\end{equation*}
the set of stationary input processes in $P _1(\overleftarrow{U}) $.

\begin{theorem}[{\bf Fixed points of $P _{\mathbf{G}}$}] 
\label{continuous foias fixed point seq}
Let $g : U \times X \to X$ be a measurable driven system with Polish input and output spaces and $(X, d_X) $ compact. Let  ${\bf  G}: \overleftarrow{U} \times \overleftarrow{X} \to \overleftarrow{X}$ be the induced driven system in sequence space  defined in \eqref{expression eq_G}. Assume now that for any element $\Theta \in P _S(\overleftarrow{U}) $ with marginal time-independent laws $\theta \in P _1(U)$, the map $g$ is a $(\theta, c _\theta)$-stochastic contraction and that one of the hypotheses {\bf (i)} or {\bf  (ii)} in Theorem \ref{Theorem_im_continuity} are satisfied for $g$.
Then, 
\begin{description}
\item [(i)] For any $\Theta \in P_S(\overleftarrow{U})$ there exists a unique $M _\Theta\in P _S(\overleftarrow{X})$ which is a fixed point of the  map $P _{\mathbf{G}}(\Theta,\cdot): P _1(\overleftarrow{X}) \to P _1(\overleftarrow{X})$ and, moreover, the map $S _{\mathbf{G}}:P _S(\overleftarrow{U})\to P _S(\overleftarrow{X}) $ that assigns $\Theta \mapsto M_\Theta$, is continuous when the domain and the image are endowed with the Wasserstein-1 distance.
\item [(ii)] When $g$ has the unique solution property and a unique measurable, causal, and time-invariant filter $U _g: \overleftarrow{U} \longrightarrow \overleftarrow{X} $ can be associated to it using \eqref{filter with USP}, we have that 
\begin{equation}
\label{filter and Foias}
S _{\mathbf{G}}(\Theta)=U _{g\, \ast} \Theta, \quad \mbox{  for all $\Theta \in P _S(\overleftarrow{U}) $.}
\end{equation}
\end{description}
\end{theorem}

\noindent\textbf{Proof.\ \ (i)} This part is proved by mimicking the proof of Theorem \ref{Theorem_im_continuity}, where the driven system $g$ is replaced by $\mathbf{G} $ and the Foias map $P _g $ by $P _{\mathbf{G}}$. In order to achieve that, we have first to show that our hypotheses on $g$ about  contractivity and uniform continuity in Theorem \ref{Theorem_im_continuity} translate into analog conditions for $\mathbf{G} $. We  recall, first of all, that since $U$ is Polish  and $X$  is Polish and compact, then so are $\overleftarrow{U}$ and $\overleftarrow{X}$ with the product topology induced by any of the metrics $d_{\overleftarrow{U}}$ and $d_{\overleftarrow{X}}$ introduced in \eqref{metric for product}. This fact allows us in particular to define the Wasserstein-1 metrics on $P _S(\overleftarrow{U}) $ and $P _1(\overleftarrow{X}) $. The hypothesis on the $(\theta, c _\theta)$-stochastic contractivity of $g$ and the stationarity of $\Theta $ imply by Propositions \ref{prop_stoch_contraction} and \ref{PG is contractive too} that $\mathbf{G} $ is a $( \Theta, c_{\theta}) $-stochastic contraction and that $P_{{\bf  G}}(\Theta, \cdot )$ is a $c_{\theta} $-contraction. Additionally, recall that by Corollary \ref{Corr_UC}, if $g$ is uniformly continuous or continuous on the first entry (as in the hypotheses in Theorem \ref{Theorem_im_continuity}) then so is $\mathbf{G} $. Given all these facts, the proof of Theorem \ref{Theorem_im_continuity} can be reproduced in our setup for $\mathbf{G} $ and $P_{\mathbf{G}}$ in order to obtain all the claims in part {\bf (i)} except for the time-stationarity of $S_{\mathbf{G}}(\Theta)$ that we postpone to the end of the proof.

\medskip

\noindent {\bf (ii)} Using the uniqueness property of the map $S _{ {\bf G}} $ that was established in part {\bf (i)}, it suffices to verify that
\begin{equation}
\label{relation with G 1}
P_{\mathbf{G}}(\Theta, U _{g\, \ast} \Theta)=U _{g\, \ast} \Theta, \quad \mbox{for any $\Theta \in P_S(\overleftarrow{U})$,}
\end{equation}
in order to prove the equality \eqref{filter and Foias}. We first recall that, as we pointed out in \eqref{fixed points and solutions}, the filter $U _g: \overleftarrow{U} \longrightarrow \overleftarrow{X} $ is the unique solution of the relation \begin{equation}
\label{ref for def of U}
T _1\circ \mathbf{G}(\mathbf{u},U _g( \mathbf{u}))=U _g( \mathbf{u}),
\quad \mbox{
for all $\mathbf{u} \in \overleftarrow{U} $}.
\end{equation}
By the definition of $\mathbf{G}  $ it is easy to see that 
\begin{equation}
\label{equiv of GG}
T _1\circ \mathbf{G}= \mathbf{G}\circ T _1,
\end{equation}
which, together with the uniqueness property in \eqref{ref for def of U} implies that $U _g $ is necessarily $T _1 $-equivariant, that is,
\begin{equation}
\label{u t equiv}
T _1\circ U _g= U _g\circ T _1.
\end{equation}
These relations imply that \eqref{ref for def of U} can be rewritten as
\begin{equation}
\label{realtion for u again}
\mathbf{G} \circ  \left(T _1 \times U _g \circ T _1\right)= U _g.
\end{equation}
This expression implies that for any time-invariant $\Theta \in P _S(\overleftarrow{U}) $ (which hence satisfies $T _{1\, \ast} \Theta= \Theta$), we have that
\begin{equation}
\label{relation with G}
\mathbf{G} _\ast \left(\Theta, U _{g\, \ast} \Theta\right)=U _{g\, \ast} \Theta.
\end{equation}
We now observe that the relation \eqref{gstar and pg are the same} that was proved in Remark \ref{gstar and pg are the same remark} for $g$ and $P _g  $ can be extended to $\mathbf{G} $ and $P _{\mathbf{G}} $. More specifically, if we consider in the space $\overleftarrow{U} \times \overleftarrow{X} $ the product measure determined by the laws of $\Theta $ and $U _{g\, \ast} \Theta $, then 
\begin{equation*}
P _{\mathbf{G}} \left(\Theta, U _{g\, \ast} \Theta\right)=\mathbf{G} _\ast \left(\Theta, U _{g\, \ast} \Theta\right)=U _{g\, \ast} \Theta,
\end{equation*}
which proves \eqref{relation with G 1}, as required. 

We conclude the proof by showing the last statement in part {\bf (i)} that was left to be proved, namely, the stationarity of $S_{\mathbf{G}}(\Theta)$. We see now that this is a consequence of the uniqueness of $S_{\mathbf{G}}(\Theta)$ as a fixed point of $P _{\mathbf{G}}(\Theta, \cdot ) $ and of the $T _1 $-equivariance of $\mathbf{G} $ in \eqref{equiv of GG}.  Indeed, on the one hand $S_{\mathbf{G}}(\Theta)$ satisfies that
\begin{equation}
\label{first fixed ptt}
P _{\mathbf{G}}(\Theta, S_{\mathbf{G}}(\Theta))=S_{\mathbf{G}}(\Theta).
\end{equation}
If we now apply $T_{1\, \ast }$ on both sides of \eqref{first fixed ptt}, use again the relation \eqref{gstar and pg are the same} for $\mathbf{G} $ and $P _{\mathbf{G}} $, and the equivariance \eqref{equiv of GG}, we  have that
\begin{multline}
\label{first fixed ptt2}
T_{1\, \ast }S_{\mathbf{G}}(\Theta)=T_{1\, \ast }P _{\mathbf{G}}(\Theta, S_{\mathbf{G}}(\Theta))=T_{1\, \ast }\mathbf{G} _\ast(\Theta, S_{\mathbf{G}}(\Theta))=\mathbf{G} _\ast(T_{1\, \ast }\Theta, T_{1\, \ast }S_{\mathbf{G}}(\Theta))\\
=\mathbf{G} _\ast(\Theta, T_{1\, \ast }S_{\mathbf{G}}(\Theta))=P _{\mathbf{G}}(\Theta, T_{1\, \ast }S_{\mathbf{G}}(\Theta)).
\end{multline}
This equality shows that $T_{1\, \ast }S_{\mathbf{G}}(\Theta) $ is also a fixed point of $P _{\mathbf{G}}(\Theta, \cdot ) $, but since that point is unique we necessarily have that $T_{1\, \ast }S_{\mathbf{G}}(\Theta) =S_{\mathbf{G}}(\Theta)$ and hence $S_{\mathbf{G}}(\Theta)  \in P _S(\overleftarrow{X}) $, as required.
\quad $\blacksquare$

\begin{example}[{\bf The unique solution of the VARMA and GARCH processes}]
\normalfont
In the examples \ref{VARMAexample} and \ref{GARCHexample} above we saw that the conditions  ${\rm E}\left[\vertiii{A(u)} \right]< 1 $ and $\alpha+\beta <1$ guarantee the stochastic contractivity of the VARMA model with time-dependent coefficients and of the GARCH(1,1) model, respectively. We saw that these conditions are vastly less restrictive than enforcing the standard contractivity of the state map that defines these models. Using now Theorem \ref{continuous foias fixed point seq} we can conclude that both models have a unique stationary solution that corresponds to the fixed points of their respective associated Foias operators. In the case of VARMA, the solution process is
\begin{equation*}
X _t=f(u _{t-1})+\sum_{k=1}^{\infty}A(u _{t-1})A(u _{t-2}) \cdots A(u _{t-k}) f(u _{t-k-1}), \quad \mbox{$t \in \Bbb Z^-$},
\end{equation*}
and for GARCH(1,1) it can be written as $r_t = \sqrt{h_t} u_{t-1}$, where
\begin{equation*}
h _t= \left\{1+\sum_{i=1}^{\infty}a(u_{t-2})\cdots a(u_{t-i-1})\right\}\omega, \quad a(u):= \alpha u ^2+ \beta, \quad t \in \Bbb Z^-.
\end{equation*}
When using the standard approach in time series analysis it is proved that these series convege almost surely (see \cite{Brandt1986}, \cite[Theorem~1.1]{Bougerol1992}, or \cite{Francq2010}). Theorem \ref{continuous foias fixed point seq}  shows that this convergence takes also place with respect to the Wasserstein distance. 
\end{example}

\section{Conclusions} 
\label{Conclusions} 

In this paper we have provided conditions that guarantee the {\bf existence} and {\bf uniqueness} of {\bf asymptotically invariant measures} for driven systems and we have proved that their dependence on the input process is {\bf continuous} when the set of input and output  processes are endowed with the Wasserstein distance.

These results have been obtained by proving the existence and uniqueness of fixed points of the associated {\bf Foias operators}, which have been profusely studied in the paper in both the state and sequence spaces. This has been achieved by using Banach's Fixed Point Theorem in the context of Foias operators by imposing readily verifiable  contractivity and continuity hypotheses that are exclusively formulated for the driven system $g$ defined in the state space. The most important condition is a 
newly introduced notion of {\bf stochastic state contractivity} for the driven system $g$, ensures that the Foias operators  in state  and in sequence spaces are also contractive with respect to the Wasserstein distance. Stochastic state contractivity is less restrictive than the standard state contractivity condition evoked to ensure the USP. 
In a future work we hope to answer more in depth the intriguing question as to how the echo state property with respect to all typical trajectories is related to the stochastic contraction property that was profusely used in this paper. 

%

\bibliographystyle{wmaainf}

\end{document}